# Finite volume method network for acceleration of unsteady computational fluid dynamics: non-reacting and reacting flows

Joongoo Jeon[1*], Juhyeong Lee[1], and Sung Joong Kim[1,2†]

[1]Department of Nuclear Engineering, Hanyang University

222 Wangsimni-ro, Seongdong-gu, Seoul 04763, Republic of Korea

[2]Institute of Nano Science & Technology, Hanyang University

222 Wangsimni-ro, Seongdong-gu, Seoul 04763, Republic of Korea

Number of pages including cover page: 53

Number of tables: 3

Number of figures: 15

Supplementary materials: 9 figures.

* Present address: Seoul National University, 1 Gwanak-ro, Gwanak-gu, Seoul 08826, Republic of Korea

† Corresponding author: Email: sungjkim@hanyang.ac.kr; Tel: +82-2-2220-2355

**Abstract**

Despite rapid improvements in the performance of central processing unit (CPU), the calculation cost of simulating chemically reacting flow using CFD remains infeasible in many cases. The application of the convolutional neural networks (CNNs) specialized in image processing in flow field prediction has been studied, but the need to develop a neural netweork design fitted for CFD is recently emerged. In this study, a neural network model introducing the finite volume method (FVM) with a unique network architecture and physics-informed loss function was developed to accelerate CFD simulations. The developed network model, considering the nature of the CFD flow field where the identical governing equations are applied to all grids, can predict the future fields with only two previous fields unlike the CNNs requiring many field images (>10,000). The performance of this baseline model was evaluated using CFD time series data from non-reacting flow and reacting flow simulation; counterflow and hydrogen flame with 20 detailed chemistries. Consequently, we demonstrated that (1) the FVM-based network architecture provided improved accuracy of multistep time series prediction compared to the previous MLP model (2) the physic-informed loss function prevented non-physical overfitting problem and ultimately reduced the error in time series prediction (3) observing the calculated residuals in an unsupervised manner could indirectly estimate the network accuracy. Additionally, under the reacting flow dataset, the computational speed of this network model was measured to be about 10 times faster than that of the CFD solver.

## 1. Introduction

### 1.1. Background

Several computational studies in various fields of research have recently investigated the reacting flows of flammable gases [1-3]. In particular, hydrogen explosion has been actively simulated using computational fluid dynamics (CFD) codes because hydrogen energy has emerged as a potential alternative to traditional carbon-based energy sources [4-6]. Hydrogen gas released from the facility can ignite depending on the mixing conditions [7]. If the ignited gas is not totally extinguished, it can lead to an explosion under geometric conditions that are conducive to flame acceleration. The prediction of this flame propagation process requires sophisticated CFD simulations [5]. [5]

Despite rapid advancements in the performance of central processing units (CPUs), the computational cost of simulating chemically reacting flow remains infeasible in many cases [5]. This is especially true for turbulent reacting flow with grid and timestep in scales less than a millimeter and a millisecond, respectively [1, 5]. These physical constraints were confirmed by recent reaction flow simulations conducted by Tolias et al. [4]. Further, they identified that approximately 10–100 h of CPU time per second of physical time was required to investigate unsteady hydrogen deflagration through CFD simulations based on a domain size of $20.0 \times 14.4 \times 12.0$ m [4]. However, in terms of their commercial application to the nuclear or hydrogen production industry, these costly simulations are impractical, given that hour-based accident analysis is required in the regulatory guide [8].

To reduce computational cost, this work focuses on the application of machine learning techniques, which are in the forefront as the most innovative technology [9]. Recently, Vinuesa et al. highlighted the role of artificial intelligence on many sectors including the energy research field [10]. As shown **Table 1**, advanced machine learning techniques have been actively applied in CFD, especially in momentum equation excluding heat transfer. Guo et al. proposed a convolutional neural networks (CNNs) approach to predict non-uniform steady laminar flow [11]. Tompson et al. suggested a CNN network for pressure solve to obtain fast and highly realistic simulations [12]. Lee et al. compared the performance of CNNs and generative adversarial network (GAN) using unsteady vortex shedding flow simulations [13]. As Vinuesa et al. noted [14], there are some differences in the modeling approach of recent machine learning studies (projecting Poisson equation, solving convective term, etc.) [15-17], but one of the main objectives of these studies is to accelerate CFD computation.

Although the application of convolutional layer in flow field prediction has been studied [18], the need to develop a unique network design fitted for CFD has recently emerged. Notably, the CNNs specialized in general image processing requires many images to train the parameters inside the kernels. For example, a recent CNN study of unsteady flow field prediction used 500,000 field images as a training dataset [13]. Takbiri et al. proposed a tier-system considering the finite volume method (FVM) and evaluated the predictive capability of this multi-layer perceptron (MLP) model for combustion simulation [19]. Stevens et al. presented a fully convolutional LSTM network where the size of the convolutional layer filter was determined considering the FVM stencil [20]. However, the stencil by diffusion term was not considered in their study because the inviscid Burger's equation dataset was used. Praditia et al. suggested a network model that adopted the numerical structure of the FVM [21]. The model solves a partial differential equation by adding a detailed network for each term of the governing equation to the framework. Although it seems intuitive, it is extremely complex, rendering it disadvantageous in terms of CFD cost and applicability to implicit numerical schemes

The aforementioned studies highlight that further network model improvement is necessary to accelerate the CFD simulation by effectively projecting the FVM in network algorithm. In addition, the performance of ML algorithms has been mainly evaluated on the non-reacting flow datasets without heat transfer (**Table 1**). For hydrogen safety, it is necessary to validate the feasibility of the network model in both non-reacting and reacting flow simulations. First, a model that can be used as baseline for comparing computational speed and accuracy must be formulated. As many studies have emphasized, the establishment of such a base model capable of performing reasonably well on a broad dataset is an essential task [22-24]. Accordingly, the objectives of this study are as follows:

1) to develop a novel concept of baseline model based on thorough comprehension of CFD and machine learning (ML) principles; network architecture and loss function

2) to evaluate the baseline model performance using non-reacting and reacting flow datasets.

*Sections 1.2 and 1.3* introduce the basic principles of FVM and neural network algorithms. *Section 2*

describes the formulation of a new baseline model. *Section 3* introduces numerical simulations for generating datasets. *Section 4* discusses the performance evaluation results of the developed model. *Section 5* investigate the effectiveness of physics-informed loss function. Finally, *Section 6* summarizes and concludes the paper.

**Table 1.** Summary of studies applying machine learning techniques to CFD simulation.

| Year | Author | Dataset type | Base network | Key ideas |
|---|---|---|---|---|
| **2016** | Guo et al. [11] | Laminar flow | CNN | Steady laminar simulation |
| **2017** | Tompson et al. [12] | Euler equation | CNN | Solving Poisson equation |
| **2018** | San et al. [25] | Wind-driven ocean circulation | MLP | Reduced order modeling |
| **2019** | Lee et al. [13] | Vortex shedding flow | GAN | Physical loss function |
| **2019** | Srinivasan et al. [26] | Low-order model of turbulent shear flow | LSTM | MLP vs LSTM |
| **2019** | Beck et al. [27] | LES closure model | RNN | CNN-based RNN architecture |
| **2019** | Takbiri et al. [19] | Gas combustion | MLP | MLP with tier system |
| **2020** | Obiols-Sales et al. [28] | Wall bounded flow | CNN | Deep learning coupled CFD for steady simulation |
| **2020** | Stevens et al. [20] | Inviscid Burgers's equation | LSTM | Convolutional layer corresponds to the stencil |
| **2020** | Ajuria et al. [29] | Incompressible flow | CNN | Solving Poisson equation |
| **2020** | Sun et al. [30] | Vascular flow | MLP | Physical constrained deep learning |
| **2021** | Kochkov et al. [31] | Incompressible N-S equation | CNN | Solving convective flux |

| 2021 | Eivazi et al. [32] | RANS equation | MLP | Solving spatial field with Physical loss function |
|---|---|---|---|---|
| **2021** | Praditia et al. [21] | Diffusion-sorption equation | MLP | Framework considering the FVM discretization |
| **2021** | Ricardo et al. [14] | Various data type | DNN | Review article |

1.2. Principles of finite volume method

The objective of this study is to develop a novel network model for accurately predicting CFD time series data. Being an application subject, a thorough understanding of the principles of CFD simulation is important. The general transport equation is given by **Eq. (1)**, where the computational domain is discretized into a finite set of control volumes (grids); $\rho$ is the density, $\phi$ is the transported quantity; $\mathbf{u}$ is the velocity vector, $\Gamma_\phi$ is the diffusion coefficient for $\phi$, $V$ is the control volume; and $\nabla\phi$ is the gradient of quantity for calculating the diffusion term.

$$\frac{\partial}{\partial t}\int_V \rho\phi \, dV + \int_V \nabla\cdot(\rho\mathbf{u}\phi)\, dV - \int_V \nabla\cdot\left(\rho\Gamma_\phi\nabla\phi\right) dV = \int_V S_\phi(\phi)\, dV \tag{1}$$

The divergence theorem, also known as Gauss's theorem, is given by **Eq. (2)**. The theorem demonstrates that the volume integral of divergence over a region is equal to the surface integral of a vector field on a closed surface (flux) [33].

$$\int_V \nabla\cdot a \, dV = \oint_A dA\cdot a \tag{2}$$

The basic FVM concept employed in most CFD codes is that divergence terms in the governing equation can be converted to surface integrals using the divergence theorem, as shown in **Eq. (3)** [33]; these terms are calculated as fluxes on each surface. The FVM is a robust conservative method because it calculates the physical quantity change by exchanging quantities among neighboring grids [34]. For example, **Eq. (4)** shows the two-dimensional (2D) axisymmetric mass continuity equation in the absence of a source term. The physical quantities of neighboring and main grids determine the physical quantity of the main grid in the succeeding timestep [34]. We focused on these FVM principles, which consider the interaction of neighboring grids and update the physical quantities with a derivative amount, for neural network modeling (more detail in *Section 2*)

$$\frac{\partial}{\partial t}\int_V \rho\phi\, dV + \oint_A (\rho u \phi)\, dA - \oint_A \left(\rho \Gamma_\phi \nabla\phi\right) dA = \int_V S_\phi(\phi)\, dV \quad (3)$$

$$\frac{\partial \rho_p}{\partial t} + \frac{\partial}{\partial x}(\rho v_x) + \frac{\partial}{\partial r}(\rho v_r) + \frac{\rho v_r}{r} = 0 \quad (4)$$

1.3. Training process of multi-layer perceptron

The techniques of ML generally aim to solve problems with complex nonlinear functions where multiple input variables are intertwined [35, 36]. This implies that multiple layers composed of multiple neurons are frequently necessary to predict the output variable based on input variables. Deep neural networks can effectively follow complex nonlinear functions by applying individual activation functions to each neuron [35, 37]. This paper explains the theoretical background for the construction of MLP based on a two-layer fully connected neural network. The two-layer network, in which $I$ denotes the dimensional input variables (feature; $X$) and $J$ is the unit number of a hidden layers (**Eq. (5)**) is shown in **Fig. 1**. The dimensions of weight vector $W$ and bias $b$ corresponding to the dimensions of input variables and hidden layers are given by **Eqs. (6)** and **(7)**, respectively. In these equations, $W^1$ is the first weight vector between input and hidden layers, and $W^2$ is the second vector between hidden and output layers. In contrast to the single neuron network, the number of parameters determined through the training process increases with the number of neurons in the hidden layer. Although increasing the number of parameters can enhance the neural network accuracy, optimizing the network size is necessary to prevent the problem of overfitting [38].

$$X \in \mathrm{R}^I, Y \in \mathrm{R}^J, Z \in \mathrm{R} \quad (5)$$

$$W^1 \in \mathrm{R}^{I\times J}, W^2 \in \mathrm{R}^J \quad (6)$$

$$b^1 \in \mathrm{R}^J, b^2 \in \mathrm{R} \quad (7)$$

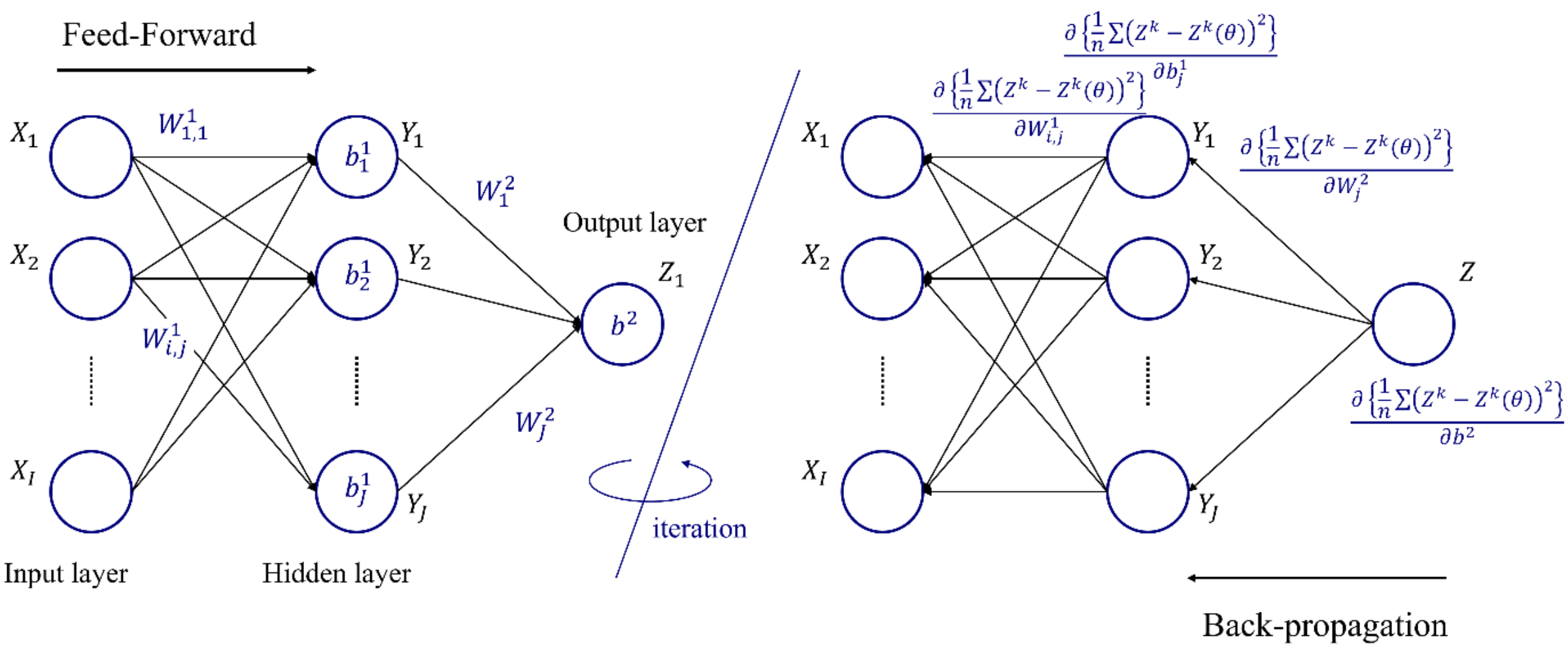

**Figure 1.** Schematic of fully connected layers. Feed-forward calculation for prediction and back propagation algorithms for parameter update.

The feed-forward algorithm proceeds as follows. The output variable is calculated using linear combinations (or nonlinear combinations with activation functions) consisting of trained parameters and input variables [39]. Each neuron in the hidden layer ($Y_j \in \mathrm{R}$) is identical to a single neuron network, as given by **Eq. (8)**, where the dimensions of each weight ($W_{i,j}^1$), input variable ($X_i$), and bias ($b_j^1$) are given by R.

$$Y_j = \sum_i W_{i,j}^1 X_i + b_j^1 \tag{8}$$

Subsequently, the values of all hidden layer neurons are correlated in the form of a linear function, as given by **Eq. (9)**. Because it is written in summation form, the matrix of variables is the same ($Z, W_j^2, Y_j, b \in \mathrm{R}$). If a nonlinear variation of the output variable exists based on input variables, then an activation function, such as ReLU, can be applied to capture the complex function, as shown in **Eq. (10)**.

$$Z = \sum_j W_j^2 Y_j + b^2 = \sum_j \left( W_j^2 \left( \sum_i W_{i,j}^1 X_i + b_j^1 \right) \right) + b^2 \tag{9}$$

$$Z = \sum_j \left( W_j^2 \cdot \mathrm{relu}(Y_j) \right) + b^2 = \sum_j \left( W_j^2 \cdot \mathrm{relu}\left( \sum_i W_{i,j}^1 X_i + b_j^1 \right) \right) + b^2 \tag{10}$$

The summation expressions of neural networks in **Eqs. (8)–(10)** can be simplified using matrix and vector notations, as shown in **Eqs. (11)–(13)**. Another important motivation for vectorization is the acceleration of training/testing calculations. To implement a neural network efficiently, vectorization leverages matrix algebra and highly optimized numerical linear algebra packages instead of using loops (summation form) to execute neural network computations rapidly [40].

$$Y = (W^1)^{\mathsf{T}} X + b^1 \tag{11}$$

$$Z = (W^2)^{\mathsf{T}} Y + b^2 = (W^2)^{\mathsf{T}} ((W^1)^{\mathsf{T}} X + b^1) + b^2 \tag{12}$$

$$Z = (W^2)^{\mathsf{T}} \cdot \mathrm{relu}(Y) + b^2 = (W^2)^{\mathsf{T}} \cdot \mathrm{relu}((W^1)^{\mathsf{T}} X + b^1) + b^2 \tag{13}$$

The prediction of a complex nonlinear function using MLP has been theoretically confirmed. However, for this to be feasible, the values of numerous parameters must be appropriately determined. Interestingly, the back-propagation algorithm allows the optimization of parameter values of each neuron in neural networks [41]. The mean square error (MSE) loss function (or cost function) for $n$ training examples $((X^{(1)}, Z^{(1)}), (X^{(2)}, Z^{(2)}),\ (X^{(k)}, Z^{(k)}), \cdots (X^{(n)}, Z^{(n)}))$ is given by **Eq. (14)**. In this equation, $Z^{(k)}$ is the ground truth value, $Z^{(k)}(\theta)$ is the value predicted by neural networks, and $\theta$ represents the set of all parameters (weight and bias).

$$J(\theta) = \frac{1}{n} \sum_{k=1}^{n} \left( Z^k - Z^k(\theta) \right)^2 \tag{14}$$

The gradient descent (GD) optimizer, which is the most representative optimizer, intuitively elucidates the parameter optimization process. With this method, parameters in MLP can be updated with each iteration, as given by **Eq. (15)**.

$$\theta \leftarrow \theta - \alpha \nabla_{\theta} J(\theta) \tag{15}$$

A more detailed parameter update procedure in a two-layer neural network is given by **Eqs. (16)** and **(17)**. The procedure shows that a one-step direct gradient calculation is difficult to implement in a layer unconnected to the output layer. For this reason, the chain rule is important to understand the back-propagation algorithm. The chain rule enables the solution of derivatives of composite functions in a considerably simple and elegant manner; it simplifies the solution of extremely complex functions [42].

$$W^2 \leftarrow W^2 - \alpha \left(\frac{\partial J}{\partial W^2}\right)^{\mathsf{T}}, b^2 \leftarrow b^2 - \alpha \frac{\partial J}{\partial b^2} \tag{16}$$

$$W^1 \leftarrow W^1 - \alpha \left(\frac{\partial J}{\partial W^1}\right)^{\mathsf{T}}, b^1 \leftarrow b^1 - \alpha \frac{\partial J}{\partial b^1} \tag{17}$$

For example, the loss function gradient of the first weight vector, $\frac{\partial J}{\partial W^1}$, can be calculated using the chain rule, as given by **Eq. (18)**. Likewise, in this study, multilayer fully connected neural networks were used to replace the nonlinear governing equations in CFD. Accordingly, the parameters were determined using the back-propagation algorithm.

$$\frac{\partial J}{\partial W^1} = \left(\frac{\partial J}{\partial Y} \cdot \frac{\partial Y}{\partial W^1}\right)^{\mathsf{T}} = \left(\frac{\partial J}{\partial Y} \cdot X^{\mathsf{T}}\right)^{\mathsf{T}} = \left(W^2 \cdot \left((W^2)^{\mathsf{T}} \cdot \mathrm{relu}\left((W^1)^{\mathsf{T}} X + b^1\right) + b^2 - Z\right) \cdot X^{\mathsf{T}}\right)^{\mathsf{T}} \tag{18}$$

## 2. Development of baseline network model

### 2.1. Formulation of FVM network

In the previous sections, the basic principles of FVM (application) and neural network algorithm (method) were presented in detail. Although recent studies have attempted to predict the CFD time series data through ML techniques [19, 43], the inherent characteristics of FVM were not sufficiently considered in network models. Neural networks with a general input/output system can facilely predict data in the training set; however, these networks have limitations in predicting the blind test set and ultimately multistep flow fields. The substantial discrepancy between ground truth data (CFD) and values predicted by ML when the distance between the parameter update and prediction time increases is a key problem that must be resolved [13]. For this reason, previous studies have validated the neural network model by limiting its application to interpolated timeline data [19]. However, this approach is not only inefficient but also infeasible for training parameters in neural networks for all computational cases. This implies that a neural network model capable of maintaining reliable accuracy even outside the training timeline must be developed. Although the CNNs specialized in image processing have been applied in flow field prediction, this image-based training approach requiring a large number of images as training examples seems inefficient in CFD simulations, where the identical governing equations are applied to all grids. Reducing the number of training samples for CFD acceleration can cause the kernel parameters to be trained incorrectly, resulting in mass and heat transfer calculation errors with the governing equations.

In this study, an FVM network (FVMN) is proposed to resolve the problems of computational cost and accuracy by considering FVM principles in the network input/output system. The FVMN architecture divides the overall process into training and prediction stages, as shown in **Fig. 2**. The TensorFlow, an open-source machine learning software, was used to develop our network model. The upper part of this figure shows that the CFD time series data at times $t$ and $(t+1)$ are used as input and output variables in the training process, respectively. It should be noted that, considering the nature of the CFD flow field where the identical governing equations are applied to all grids, the grid-based training approach was adopted in this network design. This approach bears a strong advantage over CNNs' image-based training approach in terms of accelerating unsteady CFD simulations. If the accuracy of this network design is verified, it is feasible to predict the future time series fields with only two previous fields. The parameters (weight and bias) in neural networks are iteratively updated by the Adam optimizer using the back-propagation algorithm until the loss function converges.

The basic procedure is similar to that of previous networks; however, the FVMN employs tier and derivative systems. In typical neural networks, the matrix of input and output variables can be expressed by **Eqs. (19)** and **(20)** (single-input variable condition). $Z^t$ is the output corresponding to the input at the timestep-$t$, $X^t$. Because $x_{i,j}^t$ represents the variable value at grid position $(i,j)$ at the timestep-$t$, the number of training examples is $n = M \times N$, where $M \times N$ is the number of grids in a 2D computational domain (**Eq. (21)**). The dimension of each input/output matrix due to a single input variable is $R(= R^1)$.

$$X^t = \left[x_{i,j}^t\right] \text{ where } X^t \in R \quad (19)$$

$$Z^t = \left[x_{i,j}^{t+1}\right] \text{ where } Z^t \in R \quad (20)$$

$$\left((X^{(1)}, Z^{(1)}), \cdots (X^{(n)}, Z^{(n)})\right) = \left(([x_{1,1}^t], [x_{1,1}^{t+1}]), \cdots ([x_{M,N}^t], [x_{M,N}^{t+1}])\right) \quad (21)$$

If $I$ input variables (temperature, mixture composition, etc.) are to be considered, then the matrix size is $R^I$, as follows:

$$X^t = \left[x_{1_{i,j}}^t, x_{2_{i,j}}^t, \cdots x_{I_{i,j}}^t\right]^\intercal \text{ where } X^t \in R^I \quad (22)$$

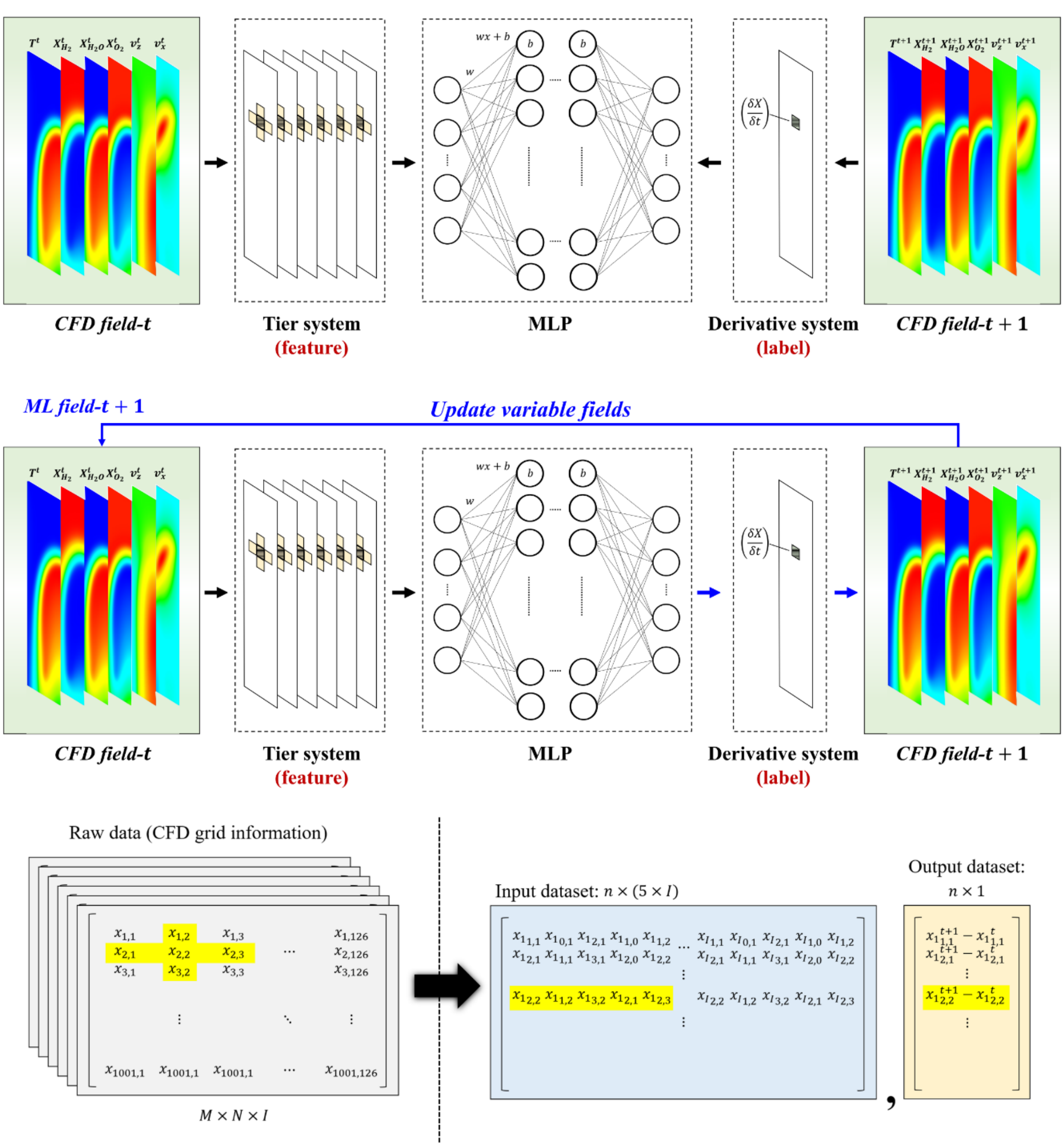

**Figure 2.** FVMN architecture with tier-derivative input/output system based on MLP. (top) training procedure; (middle) prediction procedure; (bottom) preprocessing of CFD raw data for datasets (grid-based training examples).

In contrast, in the FVMN model, the tier system was applied to the matrix of input variables. The 2D tier system employs the surrounding grids as well as the main grid as the network feature $X_t^t$. As shown in **Fig. 2** and **Eq. (23)**, a physical quantity in the main grid and neighboring grids are included in the input matrix. If $I$ variable types are considered, the matrix size is $R^{(5 \times I)}$, as given by **Eq. (24)**. This tier input system allows neural networks to imitate the mass and heat transport among neighboring cells that is considered the most important process in the FVM.

$$X_t^t = \left[x_{i,j}^t, x_{i-1,j}^t, x_{i+1,j}^t, x_{i,j-1}^t, x_{i,j+1}^t\right]^\top \text{ where } X_t^t \in R^5 \quad (23)$$

$$X_t^t = \left[x_{1_{i,j}}^t, x_{1_{i-1,j}}^t, x_{1_{i+1,j}}^t, x_{1_{i,j-1}}^t, x_{1_{i,j+1}}^t, \cdots x_{I_{i,j}}^t, x_{I_{i-1,j}}^t, x_{I_{i+1,j}}^t, x_{I_{i,j-1}}^t, x_{I_{i,j+1}}^t\right]^\top, X_t^t \in R^{(5\times I)} \quad (24)$$

Although neural networks with the tier system have been investigated in previous research, the expected enhancement in their performance remain insufficient [19]. Therefore, in this study, the derivative system which employs the change between two timesteps not the actual value of the next timestep as network output $Z_d^t$, was designed (**Eq. (25)**). We expected the enhancement of network performance through this scale separation. In fluid analysis, the scale of the actual variable value in the previous timestep is substantially larger than that in the next timestep. Because the time series data in the next timestep can be updated by simply predicting the change in the amount of $Z_d^t$, adding the actual value with a higher scale to the output variable, $Z^t = Z^{t-1} + \delta t \cdot Z_d^t$, may impair network performance. Note that several neural network studies successfully improved performance through the scale separation of output variables [44]. The concepts of derivative system and the ResNet's residual connection share similar purpose, but, at the same time, they exhibit some differences in network design. These two concepts are the same in a single variable condition. However, in multiple variables-tier system input conditions the residual connection needs to distinguish which output variable is associated with which input variable. For CFD acceleration, we expect that our network design that can be trained with a small training dataset would be advantageous. The preprocessing of CFD raw data to apply the tier-derivative system to the FVMN model is shown in **Fig. 2** (bottom). A three-dimensional raw data matrix was transformed into a two-dimensional matrix for input datasets. Because the network model was trained as an individual network for each variable, the output dataset became a one-dimensional matrix.

In summary, the neural network must be capable of predicting the change amount of each variable in the main grid $\frac{x_{i,j}^{t+1}-x_{i,j}^{t}}{\delta t}$ by the current variable values in the main/neighboring grids $X_t^t$ (**Fig. 2** (middle)). Because the network is constructed according to each output variable in this study, the dimension of the output variable matrix is $R$ regardless of the number of output variables (**Eq. (26)**); $Z_{i,d}^t$ is the output variable of the *i-th* variable. The CFD time series data at $t+1$ is finally predicted by the sum of data at $t$ and the predicted derivative amount, $X_i^{t+1} = X_i^t + \delta t \cdot Z_{i,d}^t$. $X_i^t$ is the variable field of the *i-th* variable at timestep $t$. In other words, once the network is trained with the initial CFD simulation results, it can compute a continuous time series ($X_i^{t+n} =.X_i^{t+n-1} + \delta t \cdot Z_{i,d}^{t+n-1}$).

$$Z_d^t = \left[\left(\frac{\delta x}{\delta t}\right)_{i,j}^{t+1}\right] \text{ where } Z_d^t \in R \quad (25)$$

$$Z_{i,d}^t = \left[\left(\frac{\delta x_i}{\delta t}\right)_{i,j}^{t+1}\right] \text{ where } Z_d^t \in R \quad (26)$$

The performance of ML with the general network model and FVMN model is comprehensively compared by dividing it into four cases (i.e., general, with tier system, with derivative system, and FVMN) as presented in *Section 4.2*.

## 2.2. Test matrix

The optimization of hyperparameters in neural networks has become an essential step because an overfitting problem emerges. As model complexity increases, the training and test losses generally decrease at the underfitting level of deep learning. The test dataset was not included in determining the parameters of neural networks (i.e., the training step). At a certain complexity level, the test loss starts to rebound and increase unlike the training loss. This is an overfitting problem, making it difficult to ensure model accuracy for untrained datasets. Because the FVMN aims to accurately predict the multistep CFD time series data, the optimization of neural networks is necessary to prevent overfitting.

**Table 2** lists the test matrix for optimizing the hyperparameters in the FVMN; the number of hidden layers and neurons, activation function, and learning rate. These hyperparameters to be equipped before the network training process were determined through this optimization work. Because there are several hyperparameters in neural networks, many researchers have developed systematic optimization methods such as the practical Bayesian optimization method [45] and genetic algorithm [46]. These advanced optimization methods can further increase network performance. However, the main goal of this study is to investigate the network performance according to network design in the similar network size. For this reason, the hyperparameters were determined by comparing the network performance of cases (a)–(h). As summarized in **Table 2**, the number of parameters considerably vary on a scale of 1,000–100,000 depending on the network architecture.

In cases (a)–(d), the relationship between the number of hidden layers and model accuracy was investigated. The activation function performance of ReLU and sigmoid was compared in cases (c) and (e). In cases (c), (f), and (g), the sensitivity of the number of neurons of hidden layers was investigated. Finally, in case (h), the neural network performance in the funnel-type architecture was explored based on Ref. [47]. In this study, the learning rate was fixed at 0.001 in all cases. In addition, the MSE of the validation dataset was used as a loss function to determine the parameters in the training step.

**Table 2.** Test matrix for FVMN optimization

| Case | Hidden layers | Number of parameters | Activation function | Learning rate | Loss function |
|---|---|---|---|---|---|
| **a** | 64 | 2049 | ReLU | 0.001 | MSE |
| **b** | 64, 64 | 6209 | ReLU | 0.001 | MSE |
| **c** | 64, 64, 64 | 10 369 | ReLU | 0.001 | MSE |
| **d** | 64, 64, 64, 64 | 14 529 | ReLU | 0.001 | MSE |
| **e** | 64, 64, 64 | 10 369 | Sigmoid | 0.001 | MSE |
| **f** | 128, 128, 128 | 37 121 | ReLU | 0.001 | MSE |
| **g** | 256, 256, 256 | 139 777 | ReLU | 0.001 | MSE |
| **h** | 64, 32, 16 | 4609 | ReLU | 0.001 | MSE |

## 3. Datasets

### 3.1. CFD simulation

To optimize hyperparameters of the FVMN model and evaluate the network performance, reliable CFD simulation results are required. In this study, the CFD time series data of hydrogen flame propagation were used to evaluate the network performance. Unlike flow field data generated only by the mass and momentum equations in the previous machine learning studies [11, 13, 32], the chemically reacting flow analysis includes elaborated calculations of gas species diffusion and reaction behavior. As mentioned in *Section 1*, the increase in computation time in reacting flow simulations emphasizes the necessity of accelerating calculation speed by applying ML techniques.

To clarify the efficacy of FVM principles in the network design, the flammability limit measurement simulation performed on a simple geometry with a small effect of grid skewness error was selected as the reference time series data. Despite the simple cylindrical geometry, flammability tube is a standard apparatus for hydrogen safety to measure the flammability limit according to the hydrogen mixture condition. Evaluating the network performance in the more complex apparatuses such as hydrogen burner or combiner is our future work. Recently, Jeon et al. observed the hydrogen flame extinction process and measured the hydrogen flammability limit through computational stabilized flame analysis [7]; the stabilized flame method is being vigorously applied to determine the gas flammability limit. The stabilized flame method can overcome the problem of relying on optical observations to measure the limit concentration in the standard flammability tube apparatus. With this method, suitable CFD models and grid sizes for simulating the propagation of lean hydrogen flames were identified.

Based on previous experience [7], we produced training/validation/test datasets for the FVMN by simulating a stabilized flame of 5% lean hydrogen mixture with pure air. The Fluent-CHEMKIN solver (ANSYS Fluent 18.0 solver) developed to couple the fluid dynamics with chemical reaction was employed in this study. Based on the CHEMKIN software's chemistry solving capability, the Fluent-CHEMKIN solver can work out the computational stiffness problem in reacting flow simulation by applying numerical schemes that speed up the time-to-solution [48]. The San Diego mechanism was modeled to calculate the sub-chemical kinetics of hydrogen–oxygen combustion by using a user-defined function (UDF). Although relevant studies solving the Navier-Stokes equation by using the finite element method (FEM) with high degree basic functions were reported recently [15, 49], the simulation results also can be used for training the FVMN with the appropriate grid post-processing.

The axisymmetric computational domain for the stabilized flame simulation is shown in **Fig. 3(a)**. The flame peak temperatures converged at a 0.2 mm grid size, as shown in **Fig. 3(b)**. A uniformly structured mesh (0.1 mm × 0.1 mm) was adopted in the simulation ($M = 1001, N = 126$). A coarse grid size was deemed inadequate for calculating the continuous combustion process involving the combination of gas transport and chemical kinetics. To calculate the thermal radiation for a finite number of discrete solid angles, the discrete ordinates radiation model was employed. In the gas combustion simulation, the calculation of absorption coefficient in the domain is required. For this reason, the absorption coefficient of each cell was calculated using the weighted-sum-of-gray-gases model. The model is considered as a compromise between the oversimplified gray gas model and complete model [50].

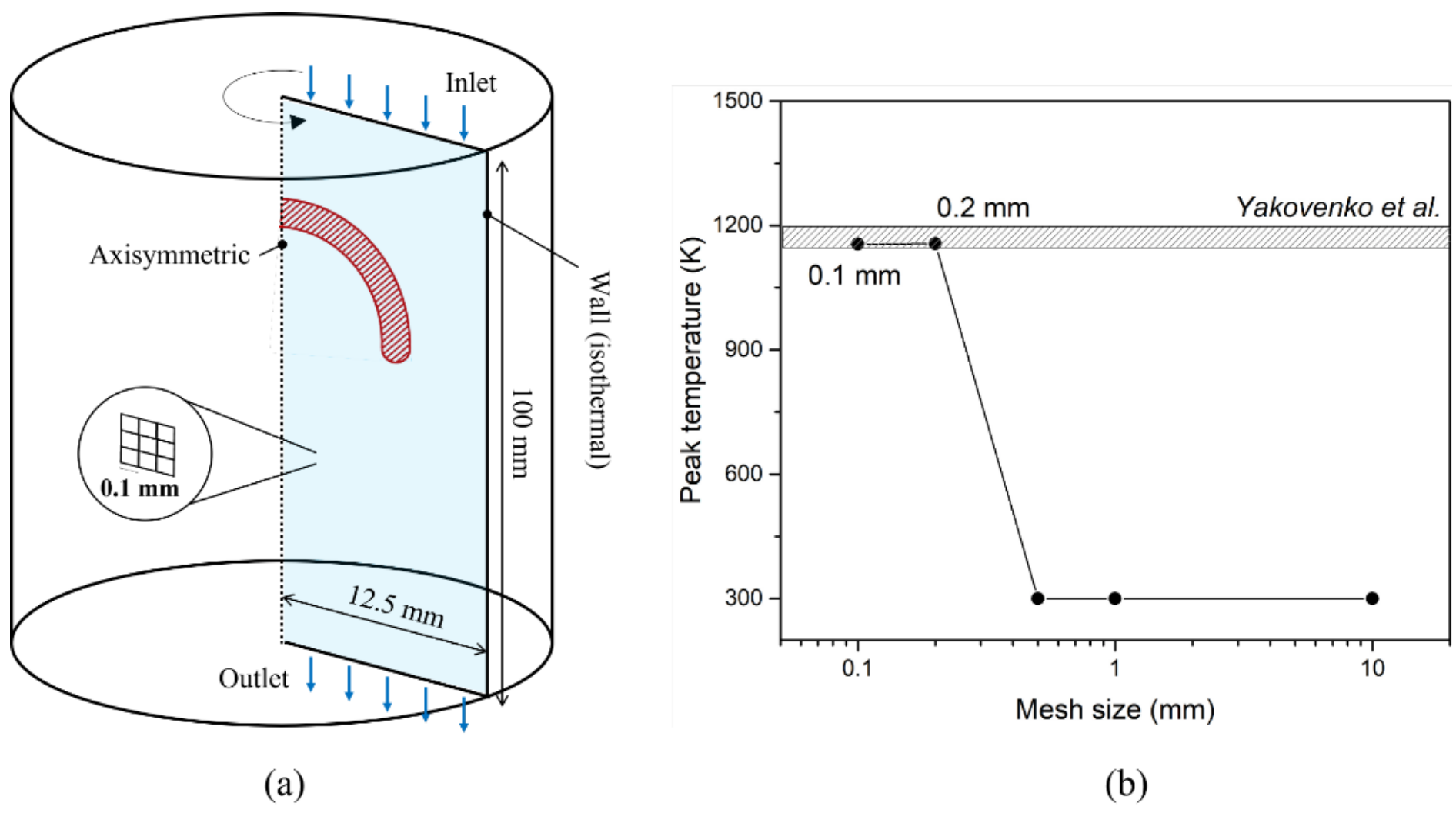

**Figure 3.** (a) Axisymmetric cylindrical domain and boundary conditions; (b) grid sensitivity analysis results based on peak flame temperature [7].

Nonslip and isothermal wall conditions were adopted because of the slight increase in wall temperature observed in the experiments [51]. The negligible effect of this slightly elevated wall temperature was confirmed by sensitivity studies [7]. Detailed descriptions of grid sensitivity and models are found in Ref. [7]. The transient solver was used because the goal of the FVMN model was to predict the CFD time series data. A 1 ms timestep was selected for the transient analysis based on a previous sensitivity study [52]. To further validate the employed unsteady flame simulation, its results are compared with those of steady simulation under the similar mixture condition shown in **Fig. S1**. The flame temperature and hydrogen mole fraction at the centerline between the two simulations showed satisfactory agreement.

Understanding the governing equation used in the current CFD simulation is essential to determine the relationship between input and output variables. The calculation process of governing equations combined with the models is described in detail in Ref. [7]. In this study, however, only representative equations confirming the nonlinearity of variables were introduced. The governing continuity and momentum equations for the unsteady solver are given by **Eqs. (27)** and **(28)**, respectively.

$$\frac{\partial \rho}{\partial t} + \nabla \cdot (\rho \vec{v}) = 0 \tag{27}$$

$$\frac{\partial}{\partial t}(\rho \vec{v}) + \nabla \cdot (\rho \vec{v}\vec{v}) = -\nabla p + \nabla \cdot (\bar{\bar{\tau}}) + \rho \vec{g} \tag{28}$$

In the foregoing equations, $\bar{\bar{\tau}}$ is the stress tensor, which can be calculated using the viscous model given by **Eq. (29)** because the calculation of the time-averaged Reynolds stress tensor under the laminar condition is unnecessary; $\mu$ is the molecular viscosity; and $I$ is the unit tensor. The nonlinearity of variables can be identified using the term $\rho\vec{v}\vec{v}$. In addition, the neighboring cell information is necessary to calculate the gas velocity gradient, $\nabla\vec{v}$.

$$\bar{\bar{\tau}} = \mu\left[(\nabla\vec{v} + \nabla\vec{v}^{T}) - \frac{2}{3}\nabla \cdot \vec{v}I\right] \tag{29}$$

In our simulation, the species transport model, which solves the conservation equation describing convection, diffusion, and detailed chemical kinetics for each component species, is utilized to calculate the flame behavior in the microregion, as given by **Eq. (30)**. To calculate the net rate of production of each species, $R_i$, the rate constants are computed using the Arrhenius equation, as given by **Eq. (31)** [53]. The San Diego mechanism was modeled to calculate the subchemical kinetics of hydrogen–oxygen combustion using a UDF. The mechanism consists of 20 reversible elementary reactions with eight reactive species including radicals; $\vec{J_i}$ is the diffusion flux of each species caused by concentration gradients, $Y_i$ (neglecting the Soret effect) (**Eq. (32)**). The diffusion flux of each species was calculated as the product of density and concentration gradients. Because density can be expressed as a function of temperature in an ideal gas model, the term leads to complex nonlinear calculations.

$$\frac{\partial}{\partial t}(\rho Y_i) + \nabla \cdot (\rho \vec{v} Y_i) = -\nabla \cdot \vec{J_i} + R_i \tag{30}$$

$$K = AT^b \exp(-\frac{E_a}{RT}) \tag{31}$$

$$\vec{J_i} = -\rho D_{i,m} \nabla Y_i \tag{32}$$

For the gas combustion analysis, the energy equation must consider reaction heat, radiation, and species diffusion. As given in **Eqs. (33)** and **(34)**, the nonlinear term is also included in the energy equation.

$$\nabla \cdot \left(\vec{v}(\rho E + p)\right) = \nabla \cdot \left(k \nabla T - \sum_i h_i \vec{J_i}\right) + S_h \tag{33}$$

$$E = h - \frac{p}{\rho} + \frac{v^2}{2} \tag{34}$$

The governing equations highlight the necessity of a MLP model to deal with the tier system and nonlinearity introduced in *Section 2*. In addition, the required input variable types for predicting time series data are confirmed. The temperature and velocity (axial and radial) variables of all grids are required to solve the continuity and momentum equations. A negligible local pressure increase was observed owing to the lean limit flame characteristics; therefore, the pressure variable was not included in the datasets. The concentration of each species is also required to calculate the diffusion flux in the energy and species conservation equations.

3.2. Datasets

The datasets for this study consist of a partial period within the entire stabilized flame simulation timeline. The entire timeline of the stabilized flame generation process, which is numerically calculated in this study, is depicted in **Fig. 4**. At 0 s (before ignition), the temperature of all grids was set to 300 K. At 0.05 s, the heat and flame propagate out of the ignition area through combustion. While the ignition energy is active, the flame continues to expand until 0.5 s. After 0.5 s, the flame begins to stabilize through heat loss balance mechanisms (conduction, radiation, and convection) and combustion heat generated by the existing flame. In ignition period (0.05-0.5 s), the ignition energy was modeled in the form of a source term on the grids at specific locations. When spatial coordinates are included in the feature, the network accuracy in the prediction dataset can decrease due to the overfitting problem. This is why the network model is more effective to aid rather than replace the unsteady CFD simulations. Consequently, the flame stabilizing period was selected as the subject of this study; hence, the simulation results for the (0.600–0.611 s) time series were used as datasets. Specifically, the results in the 0.600–0.601 s interval were used as training/validation datasets (80%/20%), and those in the 0.601–0.611 s interval were used as prediction datasets. The 0.600-0.601 s dataset consists of 0.600 s data-input and 0.600-0.601 s derivative-output.

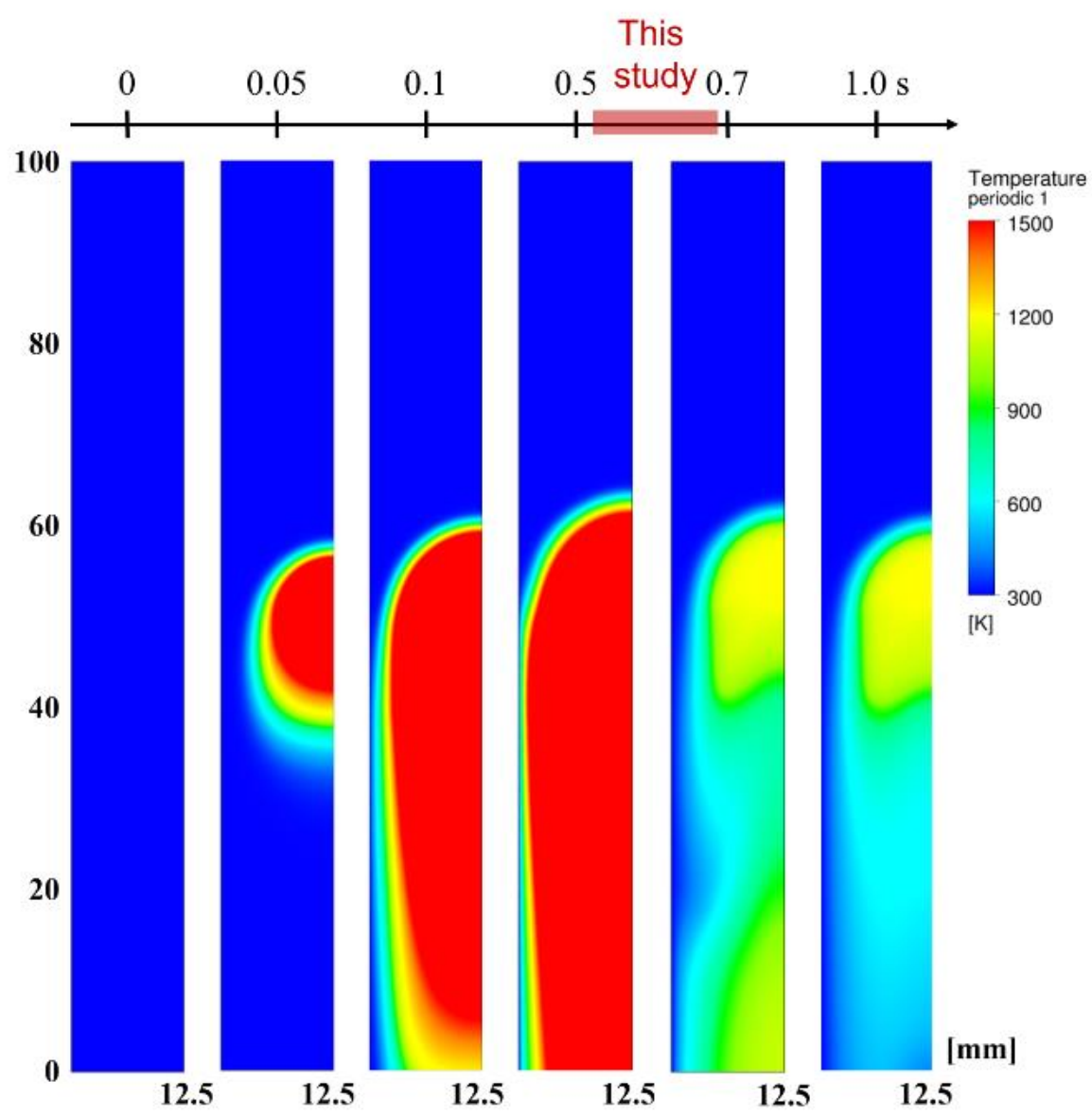

**Figure 4**. Overall stabilized flame generation process observed by CFD simulation.

As discussed in *Section 3.1*, the input variables used for machine training consist of flow variables ($v_x^t, v_r^t$), temperature variables ($T^t$), and species variables ($X_{H_2}^t, X_{H_2O}^t, X_{O_2}^t$). Unlike previous studies [43], the grid location information was not included in input variables to confirm the universal feasibility of the FVMN model in multistep time series. The concentrations of radicals, such as H and O, were not included in input/output variables because they were scaled close to zero, even in the reaction zone.

Note that investigating the effect of including radical variables on input variables is beyond the scope of this work. This will be considered in future work.

In data-driven research, the implementation of feature scaling to render all features similar in standing is necessary; hence, the effect of a specific feature cannot cause bias. The original input variables exhibit a large-scale difference according to their type, as shown in **Fig. S2(a)**. Specifically, the scale difference between temperature at a scale of $10^3$ and hydrogen mass fraction at a scale of $10^{-3}$ was significant. For this reason, the standardization of input variables, which is a representative scaling technique where the values are centered about the mean with a unit standard deviation (**Eq. (35)**), is used in this study. The ranges of standardized input variables are confirmed to be considerably similar, as shown in **Fig. S2(b)**. The two-dimensional standardized data distribution in the training/validation examples based on temperature and axial velocity is shown in **Fig. S3**.

$$x^t = \frac{x_{raw}^t - \mu}{\sigma} \tag{35}$$

Six distinct networks were trained to predict each derivative amount of output variable $\left(\frac{v_x^{t+1}-v_x^t}{\delta t}, \frac{v_r^{t+1}-v_r^t}{\delta t}, \frac{X_{H_2}^{t+1}-X_{H_2}^t}{\delta t}, \frac{X_{H_2O}^{t+1}-X_{H_2O}^t}{\delta t}, \frac{X_{O_2}^{t+1}-X_{O_2}^t}{\delta t}, \frac{T^{t+1}-T^t}{\delta t}\right)$. In other words, the data series in timestep $t$ is used as input, and that in timestep $t+1$ is used as output during the training process, as shown in **Fig. 2**. More specifically, the raw data matrix from CFD data is transformed into the matrix of input and output variables through the tier and derivative systems, respectively, as discussed in *Section 2.3*.

### 3.3. ML domain

As shown in **Fig. 5**, the computational domain is numerically solved by dividing it into ML and CFD domains. In the CFD domain, the flow field in the succeeding timestep is calculated by solving the first principles (i.e., governing equations presented in *Section 3.1*). In the ML domain, the time series data are calculated using the FVMN model only. The dimensions of the original data matrix in the entire domain, $A$ (before data preprocessing for ML), is $\mathrm{R}^{M,N}$, where $M$ and $N$ are the number of grids in the axial and radial directions, respectively. The domain for $M^*$ layers in the inlet and outlet regions, $A_{inlet}$ and $A_{outlet}$, respectively, are classified in the CFD domain ($M^* = 100$), as given by **Eq. (36)**. The rest of the domain, excluding the CFD domain (inlet/outlet regions), was designated as the ML domain, $A_{flame} \in R^{M-2M^*,N}$. In other words, the input and output variables for the FVMN are extracted from the ML domain, $A_{flame}$, as shown in **Eq. (37)**. This is equivalent to expressing the number of training examples as $n = 801 \times 126$ (grid-based training examples).

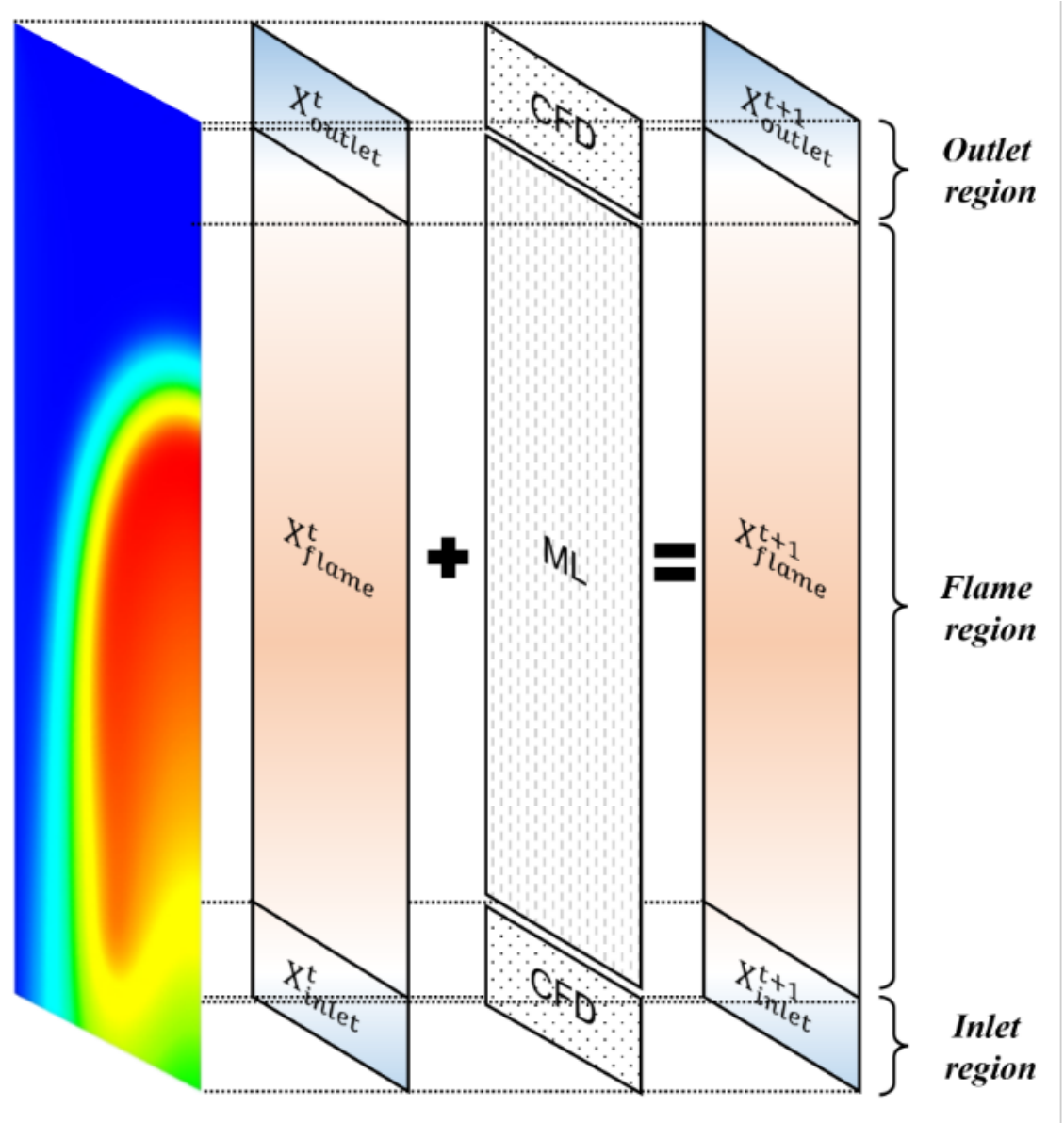

**Figure 5**. Schematic of CFD-coupled ML method; computational domain is divided into three regions, of which input/output and flame regions are updated by CFD and ML, respectively.

The use of the combined ML–CFD calculation method in this study is suggested for two reasons. First, the height of the selected computational domain in this study is 100 mm, indicating that the distance between the main flame calculation region and inlet/outlet is extremely short. For this reason, the variation in the time series data of the inlet and outlet regions exhibits stiffness in the numerical solution. In the preliminary analysis, this stiffness can significantly limit the application of data-driven methods. Second, as introduced in *Section 1*, the sharp decrease in accuracy is identified as the timestep in many ML–CFD application studies. These limitations may be caused by the inability of the neural network to imitate the boundary conditions of CFD simulation. The combined calculation method is implemented to investigate whether it can alleviate the intrinsic limitations of data-driven methods.

$$A^t \in \mathrm{R}^{M,N},\ A_{inlet}^t \in R^{M^*,N}, A_{outlet}^t \in R^{M^*,N}, A_{flame}^t \in R^{M-2M^*,N} \tag{36}$$

$$X_t^t \subset A_{flame}^t, Z_d^t \subset A_{flame}^{t+1} \text{ where } X_t^t \in R^{(5\times I)}, Z_d^t \in R \tag{37}$$

3.4. Boundary conditions

As discussed in *Section 2.3*, the tier system, which includes information on neighboring grids as input variables, is applied to the FVMN model. The walls in the CFD simulation (**Fig. S4**), except for the inlet and outlet, have remaining boundary conditions. For the center domain, where the main grid is located at the second layer or beyond, input variables can be determined in the normal tier system, as given by **Eq. (38)**. In contrast, the boundary layer (first grid layer) lacks grid information beyond the wall. If the dimension of the input variable matrix varies according to the grid position, then the training loss is expected to increase.

In the axisymmetric wall, this component shortage problem can be resolved using a technique similar to the method for solving the zero Neumann boundary condition in the finite difference method. As shown in **Fig. S4** and given by **Eq. (39)**, the variable information of the main grid can replace that of the grid beyond the wall. Because a rest wall is under isothermal conditions, wall information is required, as indicated by **Eq. (40)**. To apply wall information, the temperature variable should be set as wall temperature, and other variables (species concentrations as well as axial and radial velocities) are set as zero, that is, $x_{1,wall} = T_{wall}, x_{2,wall} = x_{3,wall} = x_{4,wall} = x_{5,wall} = x_{6,wall} = 0$ . However, the application of wall condition is observed to induce a higher error than the zero Neumann boundary condition, as shown in **Fig. S5**. When wall information was included in the wall boundary domain, a discrepancy between the scales of input values and major domain values emerged, magnifying the training error. Because the training examples containing wall information constitute a small portion of the entire sample, distinguishing this dataset during training is seemingly difficult for neural networks. Accordingly, in this study, the component shortage problem in the wall is resolved using the condition of symmetry (**Eq. (41)**).

The prediction of succeeding CFD time series data using the FVMN results from the standardization of input variables, domain partitioning, and zero Neumann boundary condition, which is described in this section; the results are presented in *Section 4*.

$$(Z_d^t)_{center} = f\left\{\left(x_{k_{i,j}}^t, x_{k_{i-1,j}}^t, x_{k_{i+1,j}}^t, x_{k_{i,j-1}}^t, x_{k_{i,j+1}}^t\right)\right\}_{k=1}^{6} \quad \text{where } x_k^t \in X_t^t \tag{38}$$

$$(Z_d^t)_{axis} = f\left\{\left(x_{k_{i,j}}^t, x_{k_{i-1,j}}^t, x_{k_{i+1,j}}^t, x_{k_{i,j-1}}^t, \boldsymbol{x}_{\boldsymbol{k}_{\boldsymbol{i,j}}}^{\boldsymbol{t}}\right)\right\}_{k=1}^{6} \quad \text{where } x_k^t \in X_t^t \tag{39}$$

$$(Z_d^t)_{wall} = f\left\{\left(x_{k_{i,j}}^t, x_{k_{i-1,j}}^t, x_{k_{i+1,j}}^t, \boldsymbol{x}_{\boldsymbol{k}_{\boldsymbol{wall}}}^{\boldsymbol{t}}, x_{k_{i,j+1}}^t\right)\right\}_{k=1}^{6} \quad \text{where } x_k^t \in X_t^t \tag{40}$$

$$(Z_d^t)_{wall} = f\left\{\left(x_{k_{i,j}}^t, x_{k_{i-1,j}}^t, x_{k_{i+1,j}}^t, \boldsymbol{x}_{\boldsymbol{k}_{\boldsymbol{i,j}}}^{\boldsymbol{t}}, x_{k_{i,j+1}}^t\right)\right\}_{k=1}^{6} \quad \text{where } x_k^t \in X_t^t \tag{41}$$

## 4. Evaluation of network performance

### 4.1. Optimization of hyperparameters

As explained in *Section 2.2*, the degradation of neural network performance despite increasing network size is a typical problem occurred in ML methods (overfitting problem). For this reason, optimizing the most important hyperparameters such as the number of layers and nodes is necessary for the baseline model. In this study, the hyperparameters were optimized based on the errors of trained FVMN in the training/validation datasets. More specifically, the network model was trained with 80% grid of the (0.600–0.601 s) time series data, as shown in **Fig. 2** (top), where the number of total grids is 100,926. 80% of the total grids was 80,741 and 20% was 20,185. The MSE for the rest 20% data (validation dataset) was calculated for each epoch, and the training process was terminated when the loss function was converged (**Fig. S6**). By the loss variation trend, it was confirmed that the amount of data was sufficient to prevent overfitting. After training, the trained networks predicted the 0.601 s data using the 0.600 s data as input, as shown in **Fig. 2** (middle). **Table 2** summarizes the network structure in each case.

The error distribution (relative error of each grid) between ground truth (CFD) and predicted values based on the temperature field at 0.601 s is shown in **Fig. 6**. The color scale is the same for all cases (0–0.05%). Case (a), which had the simplest network structure, exhibited considerable errors in most of the interface region between flame and unburned gas owing to the underfitted networks. In cases (c) and (d), the underfitting problem was gradually alleviated with an increasing number of hidden layers. However, when the number of hidden layers increased to four, case (d) had errors higher than those in case (c). The main error regions were the flame concave region and outlet region where variable change with time was noticeable. This implies that the overfitted networks cause a considerable number of errors in the stiff calculation region.

This overfitting phenomenon is also confirmed to occur with increasing number of neurons per hidden layer, as shown in cases (f) and (g). Although training of parameters more than 10 times is required in case (g) compared with case (c), the errors in the concave region are higher. The funnel-shaped structure, which has been widely used in the field of image learning, is unable to achieve noticeable improvements in accuracy. In case (e), using the sigmoid activation function, the maximum and mean relative errors were higher than those when the ReLU function was employed. In addition to the comprehensive comparison based on error distribution, the maximum and mean relative errors in each case are quantitatively compared in **Fig. S7**. In both error graphs, the underfitting and overfitting levels can be clearly identified. Accordingly, the FVMN was optimized in the structure with three hidden layers and 64 nodes per hidden layer (case (c)). The structural optimization results were also verified for prediction datasets (i.e., 0.602–0.611 s data). The following results were obtained based on the optimized network structure.

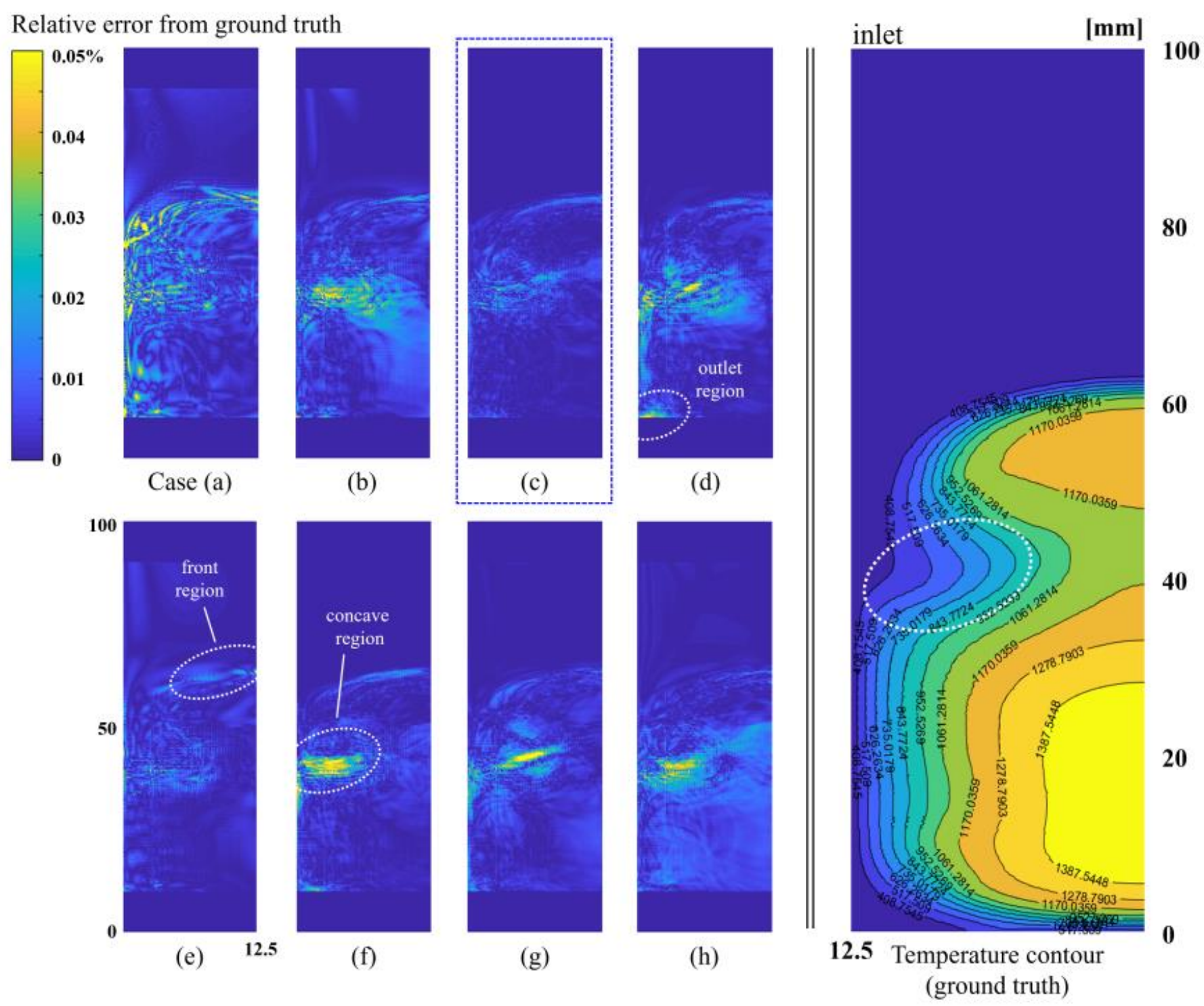

**Figure 6**. Relative error distribution of predicted temperature, $\frac{|T_{ML}-T_{CFD}|}{T_{CFD}}$, based on CFD results at 0.601 s; **Table 2** lists properties of each network model.

4.2. Performance of FVMN compared with general network model

The novel concept in the FVMN model that embodies CFD principles has the advantage of being intuitive; whether this concept can actually improve performance must be investigated. The effects of using the tier and derivative systems on prediction accuracy based on the temperature value at 0.601 s are shown in **Fig. 7**. The input data of all network models including the FVM network model and the general network model were standardized using the Python "describe()" method (**Eq. (35)**). In the case of output data, no difference in performance was identified according to standardization or not. For all cases, the hyperparameters, including network size, are identical to the optimized parameters presented in *Section 4.1*. Cases (1)–(4) only differ in terms of the form of input and output variables during training and prediction processes. As discussed in *Section 2.3*, in case (1) with the FVMN, the input and output variable matrices are preprocessed by the tier and derivative systems, as given by **Eqs. (42)** and **(43)**, respectively:

$$X_t^t = \left[\left\{\left(x_{k_{i,j}}^t, x_{k_{i-1,j}}^t, x_{k_{i+1,j}}^t, x_{k_{i,j-1}}^t, x_{k_{i,j+1}}^t\right)\right\}_{k=1}^{I}\right]^{\mathsf{T}} \; X_t^t \in R^{(5\times I)} \tag{42}$$

$$Z_{i,d}^t = \left[\left(\frac{\delta x_i}{\delta t}\right)_{i,j}^{t+1}\right] \text{ where } Z_d^{t+1} \in R \tag{43}$$

For case (2) with the general network model, only the actual variable information of the main grid was used as input and output variables, as given by **Eqs. (44)** and **(45)**, respectively:

$$X^t = \left[x_{1_{i,j}}^t, \cdots x_{I_{i,j}}^t\right]^{\mathsf{T}}, X^t \in R^I \tag{44}$$

$$Z_i^t = \left[x_{i_{i,j}}^{t+1}\right] \text{ where } Z^t \in R \tag{45}$$

The effects of tier and derivative systems were investigated in cases (3) and (4). The network model using only the tier system is given by **Eqs. (46)** and **(47)**, whereas **Eqs. (48)** and **(49)** give the model with only the derivative system.

$$X_t^t = \left[\left\{\left(x_{k_{i,j}}^t, x_{k_{i-1,j}}^t, x_{k_{i+1,j}}^t, x_{k_{i,j-1}}^t, x_{k_{i,j+1}}^t\right)\right\}_{k=1}^{I}\right]^{\mathsf{T}} \; X_t^t \in R^{(5\times I)} \tag{46}$$

$$Z_i^t = \left[x_{i_{i,j}}^{t+1}\right] \text{ where } Z^t \in R \tag{47}$$

$$X^t = \left[x_{1_{i,j}}^t, \cdots x_{I_{i,j}}^t\right]^{\mathsf{T}}, X^t \in R^I \tag{48}$$

$$Z_{i,d}^t = \left[\left(\frac{\delta x_i}{\delta t}\right)_{i,j}^{t+1}\right] \text{ where } Z_d^t \in R \tag{49}$$

With the foregoing, significant improvement in the performance of the FVMN approach compared with the general model is confirmed, as shown in **Fig. 7**. In case (2), the relative error with respect to CFD data exceeded 0.05% not only in the interface region between the flame and unburned gas but also in virtually all regions with burned gas. The maximum relative error was 1.12%, which was substantially higher than that of the FVMN model (0.04%). Given that CFD simulation calculations characteristically require many timesteps, this error level in a single step is deemed unacceptable. In case (3), the errors in some grids were alleviated by applying the tier system (maximum relative error: 0.26%).

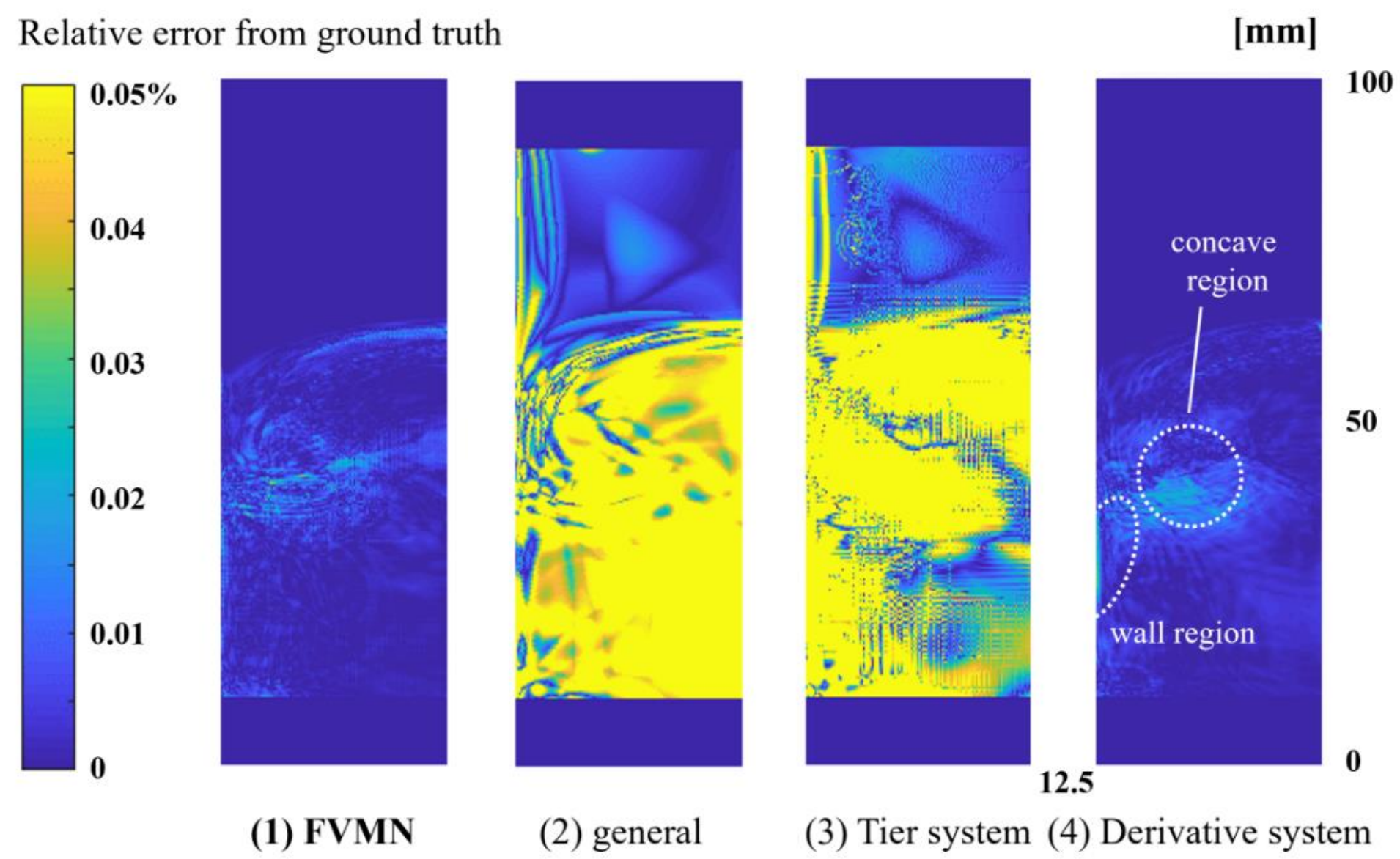

**Figure 7**. Relative error distribution of predicted temperature, $\frac{|T_{ML}-T_{CFD}|}{T_{CFD}}$, depending on application of tier and derivative systems (based on CFD results at 0.601 s).

Interestingly, the efficacy of the derivative system is considerably more distinct than that of the tier system, although previous studies have only emphasized the importance of the latter [19, 43]. Case (4) demonstrates that the application of the derivative system significantly reduces errors even in the stiff calculation region. As discussed in *Section 2.3*, the scale of the actual variable value of the previous timestep is substantially larger than that of the next timestep. For example, the scale of temperature values reached approximately 100 times the scale of temperature change (in a single timestep) in our CFD datasets. Accordingly, the scale separation of output variables by the derivative system is concluded to have a significant effect on improving network performance. These results highlight that the derivative-output system is essential to the utilization of data-driven method in CFD applications.

Although the overall error distribution between cases (1) and (4) is similar, the local errors at the wall boundary and flame concave regions are higher in case (4). If only the main grid information is included with input variables without the tier system, identifying the wall is expected to be difficult for network grids. Because the dimensions of the problem also increase as the number of input variables increases, the effect of the tier system may be more distinct on larger training datasets. In conclusion, the FVMN concept, which embodies the principles of the CFD calculation procedure, affords the clear advantage of increasing the accuracy of time series data prediction.

### 4.3. Performance of FVMN in multistep prediction

The ultimate goal of FVMN is to reduce the cost of reacting flow simulations by accelerating CFD computations. To achieve this, the trained networks must be capable of precisely predicting multistep CFD data that are not included in the training dataset. In this study, the FVMN performance in multistep prediction using our CFD data (*Section 3.1*) is evaluated. The single-step time series data (0.600–0.601 s) derived from the CFD simulation were used as the training dataset, and multistep data (0.601 – 0.611 s) were utilized as prediction datasets (i.e., test datasets). If the multistep training dataset is used, it may be possible to maintain network performance over a broader time range; however, a larger network size and longer training time are required. Note that the investigation of the change in network performance according to the training set size is beyond the scope of this work.

The variation in the performance of FVMN according to the timestep progress (blue rectangular line) is shown in **Fig. 8**. The maximum relative error (filled symbol) and mean relative error (hollow symbol) virtually show the same trend. For the trained data at 0.601 s, the identified maximum relative error was 0.04%, exhibiting an extremely small mean relative error (0.002%). In the first prediction dataset (at 0.602 s), the maximum and mean errors were 0.10% and 0.006%, respectively. Based on the prediction dataset, the output variables were calculated as the sum of ML time series data in the previous timestep and the derivative predicted by ML data, $X_{i,ML}^{t+1} = X_{i,ML}^{t} + \delta t \cdot Z_{i,d}^{t}(X_{t,ML}^{t})$. In other words, in the ML domain, the network model was completely independent of CFD data (ground truth). In the next prediction dataset (at 0.603 s), the maximum and mean errors were 0.21% and 0.015%, respectively. Both errors in prediction datasets increased in the form of a quadradic function for the timestep. In the last dataset (0.611 s), the maximum error was approximately 2.6%. This is in contrast with the error exceeding 2% in the second timestep in the general network model (diamond symbol). Considering that this case study is a complex reacting flow simulation that includes 20 detailed chemical reactions, a mean error of less than 0.3% (for the prediction dataset, which is 10 times the size of the training dataset) indicates reliable network performance.

The FVMN computational time was compared with the time consumed by the CFD solver. In the CFD simulation, an unsteady time series is calculated by computing the flux balance of each governing equation while reducing the residual through internal iteration. Therefore, reacting flow simulations accompanied by convoluted non-linear governing equations requires very long calculation time. On the contrary, the neural networks approximating the non-linear equations by combining linear functions and activation functions together can compute next time series relatively quickly. Although neural network performance was more advantageous for GPU servers, an Intel Core i7-6700 CPU (TensorFlow 2.3.0) was used in both calculations for direct comparison. The average time required to calculate the succeeding timestep data was confirmed approximately 10 times faster in the FVMN ($t_{CFD}$~60 $s$, $t_{ML}$~5$s$). The CFD computation time is the average time from 0.600-0.611 s and the network computation time was measured using time module in Python. This reduction in computational cost will be more manifest when a GPU server is used. Under the current grid number condition, the GPU system showed approximately 10 times speedup over the CPU system for $t_{ML}$. Interestingly, it was confirmed that the speedup rate increased with the grid number. In this experiment, the utilized GPU system were NVIDIA TESLA A100 (CUDA core, 40GB MEM). Although the initial network training time should be considered for the comprehensive acceleration evaluation, the training time cannot have a significant effect on the long-term CFD simulation.

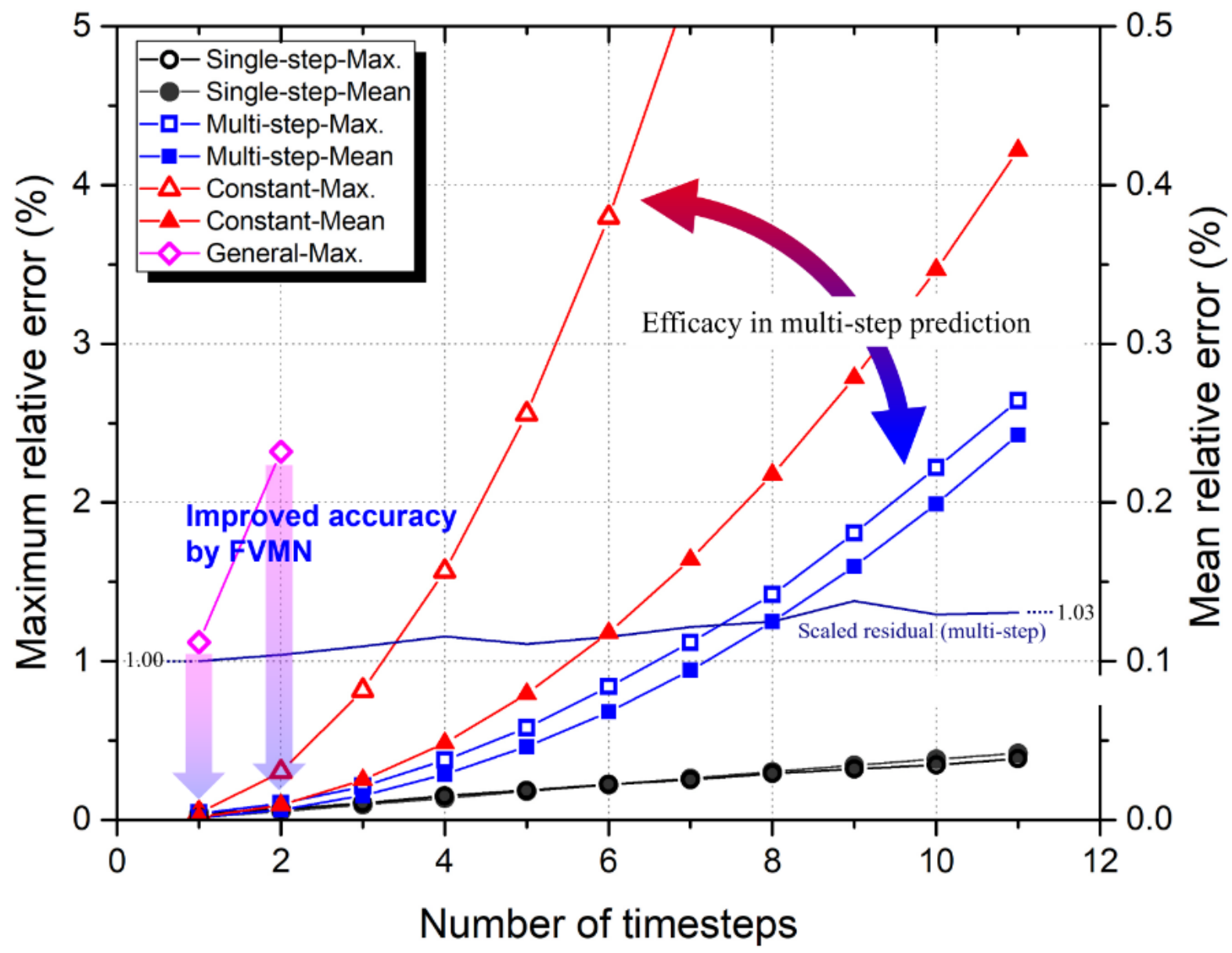

**Figure 8**. Changes in maximum and mean relative errors in single-step prediction, multistep prediction, and constant gradient cases; feasibility of multistep prediction with FVMN is confirmed.

Two error categories can account for the discrepancy between ground truth and ML time series data: initial condition error and gradient error. Because the output variables were calculated based on previous ML data instead of CFD data, the errors generated in previous calculations were already reflected in the initial condition. In addition, an error is generated by the network in predicting the gradient using input variables. In ascertaining the impact of each error, these errors were observed in the single-step prediction case. The single-step prediction case means that the derivative in each timestep is calculated using the CFD time series data, $X_{i,ML}^{t+1} = X_{i,CFD}^{t} + \delta t \cdot Z_{i,d}^{t}(X_{t,CFD}^{t})$. Removing the initial condition error implies that only the magnitude of the gradient error requires investigation. As shown by the black circular lines, the maximum and mean gradient errors linearly increase with time. In this case, the maximum error did not exceed 0.5%, even at the last timestep. This confirmed that the error accumulation in multistep prediction was mainly caused by initial condition errors. The linearly increasing error observed in the single-prediction case was an accumulation of errors, resulting in a quadratic error function in the multistep prediction case. If the gradient error is reduced by further optimizing the FVMN or advancing the loss function, then accelerating the CFD calculation over a wider time range is feasible. However, this linear error increase is considered inevitable with an increasing gap in the training and prediction points. The solution to this linear increase in the gradient error is discussed in *Section 5.3*.

In addition, the red triangular lines represent the accumulated error when the gradient of each variable obtained from the data at 0.600–0.601 s was constantly applied to the entire test dataset. Although a small size of each timestep (0.001 s) and a small number of timesteps (10 times) were applied, the error increase in the cases in which the FVMN was used was considerably fast. In the 4$^{th}$ prediction dataset, the maximum error exceeded 2%. These results highlight the necessity of sequential gradient prediction using ML to accelerate the CFD simulation.

The relative error distribution corresponding to each timestep is shown in **Fig. 9**. The main reason for the gradient error, which was difficult to ascertain through maximum error analysis, can be inferred from the error distribution in the ML domain. As discussed above, in the multistep prediction case (**Fig. 9(a)**), the overall error gradually increased over time. The figure depicts the flame front with the error distribution in the error field at 0.604 s. The local error was confirmed to generally increase around the flame front, especially in the region near the concave front. The error generated in the concave region propagated to its vicinity as the ML prediction proceeded. Although the magnitude of the relative error exhibits some variations, this error propagation pattern is considerably similar to the single-step prediction case, as shown in **Fig. 9(b)**.

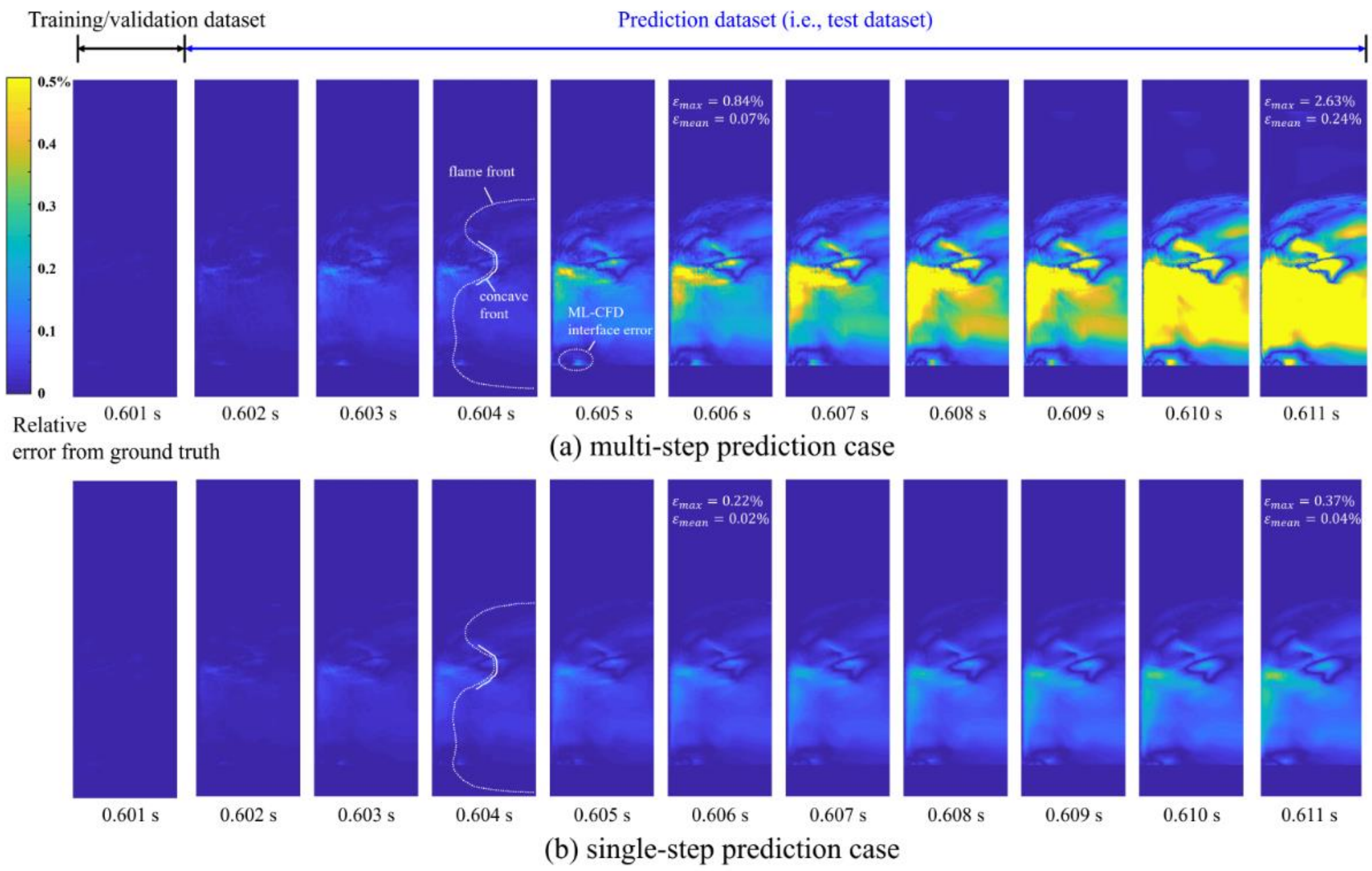

**Figure 9**. Relative error distribution of predicted temperature $\frac{|T_{ML}-T_{CFD}|}{T_{CFD}}$ over a timestep. Error accumulates in multistep case unlike in single case where error linearly increases.

**Fig. 10** shows the accuracy of the FVMN model by comparing the ground truth and predicted values of each variable at 0.611 s. The axial profiles near each wall (isothermal wall and axisymmetric wall) and at the center were used for validation. At this time, the variables near each wall were compared based on the first grid from the wall. In the case of axial velocity, the fluctuation of the flow decreases as it approaches the isothermal wall. The intensity of the upward flow is greatest at the axisymmetric wall, which is the center of the flow. In all axial profiles of the hydrogen mass fraction, the value decreases and reaches zero as it passes through the unburned gas section.

It was noted that the network provided fairly accurate solutions for almost all domains. No noticeable errors were identified in all axial variable profiles except for one point of hydrogen fraction at 10 mm of the isothermal wall. The grid where this divergence occurred was both the boundary of the CFD and ML domains (detailed in *Section 3.3*) and the boundary with the isothermal wall. The methodology of the input data preprocessing for this double boundary layer is our future work. Nevertheless, it was verified that the FVMN model showed satisfactory agreement with the ground truth if the appropriate data preprocessing at the boundary layer was accompanied.

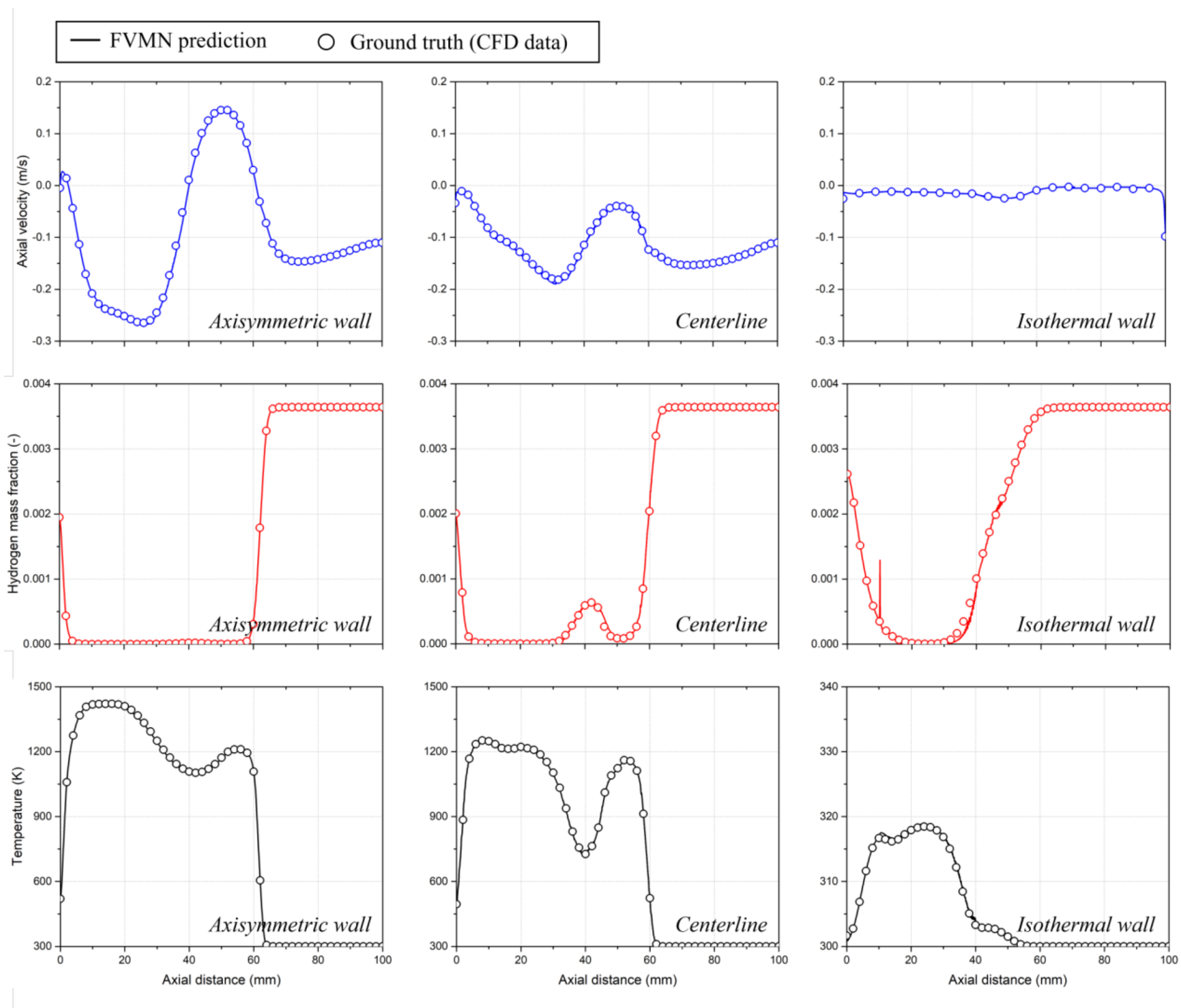

**Figure 10**. Comparison of axial variable profile between FVMN results and ground truth (CFD data) at 0.611 s (top: axial velocity, middle: hydrogen mass fraction, bottom: temperature).

The first reason for the noticeable error in the concave region is its gradient size being larger than those in other regions. As the predicted gradient increases, the relative error with respect to the actual value becomes increasingly disadvantageous. The second reason is possibly the limited ratio of concave region data included in the training dataset. The distribution of output variables, $Z_d^{t+1}$, based on the training/validation dataset is shown in **Fig. S8**. The distribution has a clear deviation from the normal distribution which is generally preferred for improving the ML training effectiveness. The biased dataset distribution contains output variables ranging from $-2.5\ to\ 2.5\ K$; this is approximately 60% of the total data. However, the concave region, where a relatively large error occurs, includes a higher $\Delta t$ region, as shown in **Fig. S8**. Consequently, the network performance in a specific region can be degraded when the network is not sufficiently trained to use partial data with a relatively small distribution. The error near the wall identified in **Fig. 10** seems to have increased for the same reason. The gradient error may be reduced if an advanced sampling method is used in the training process. Because the aim of this study is to develop a baseline FVMN model, this aspect must be investigated in future work.

## 5. Physic- informed loss function

In *Section 4*, the neural network performance improvement by the tier-input and derivative-output system was identified. The results demonstrated that projecting the FVM principle on the network structure can increase the prediction accuracy of CFD simulation of heat transfer and flow field. However, the numerical algorithm of FVM for handling governing equations (nonlinear partial differential equations) is difficult to be implemented sufficiently into the network design only by the proposed input/output system. In this section, we introduced a physics-informed loss function, which calculates balance of each governing equation by summation of physical fluxes, in the network to further assimilate FVMN to FVM. Recently, Hamidreza et al. showed that the steady turbulent flow field can be successfully predicted with a customized loss function, which is summation of the supervised loss and residual of the governing equations [32]. Although recent studies have investigated the effects of physics-informed loss functions on machine learning algorithms [32, 54], relevant studies predicting uncalculated future time series of unsteady CFD simulation are not reported in sufficient detail.

Because the effect of the loss function on the network performance can be attenuated in the reacting flow simulation condition which included species and energy balance equations, a counterflow simulation including only mass and momentum balance equations was utilized in this section. It is noteworthy that CFD analysis of flow field before the start of the reaction is also important for hydrogen safety. The TensorFlow provides the “tf.GradientTape” application programming interface (API) for automatic differentiation as part of the back-propagation algorithm [32].

5.1. Laminar flow dataset

To investigate the efficacy of the physics-informed loss function, a 2D incompressible counterflow dataset was generated by the OpenFOAM code. The OpenFOAM solver is icoFoam, which is transient solver for incompressible laminar flow of Newtonian fluids (**Eqs. (50) and (51)**). The geometry of the simulation is a simple rectangle as shown in **Fig. 11(left)**. The left and right wall has 1 m height, and top and bottom sides have 2 m width. The 0.2 m scale inlets are specified in the middle of the left and right walls, and the top and bottom walls are defined as the outlets. The rest of the boundaries was set as walls. The mesh was constructed by 20,000 hexahedral cells. Through the overall same mesh quality, the maximum non-orthogonality was 0 and maximum skewness was 6.66e-14. The inlet 1 located at the middle of left wall was assigned to 2 m/s of positive x-axis velocity ($u$) and 1 m/s of negative x-axis velocity in inlet 2 (nonsymmetric inlet condition). As the working fluid, a virtual fluid with a kinematic viscosity of $0.01m^2/s$ was modelled. The simulation time was 4 seconds, and each timestep was equally controlled as 0.01 s.

$$\nabla \cdot u = 0 \tag{50}$$

$$\frac{d(u)}{dt} + \nabla \cdot (u \otimes u) - \nabla \cdot (\nu \nabla u) = -\nabla p \tag{51}$$

**Fig. 11(right)** shows pressure and velocity fields according to time transient; p represents pressure field, u and v each represents the x-velocity and y-velocity. Because the velocity of inlet 1 is twice than velocity of inlet 2, this counterflow showed biased results from left side to the right side and y-velocity shows symmetric value along central axis ($y = 0$). In 2 seconds the pressure field showed more biased value to the right side and counterflow of x-axis induces more y velocity gradient. In 3 seconds, the velocity and pressure fields were stabilized near the inlet 2 and then simulation proceeded without significant change after this time. The 1.00-1.11 s timeseries dataset, which is one of the most unsteady phases, was used for network performance evaluation. The input and output variables for the FVMN were extracted from the ML domain ($n = 180 \times 100$).

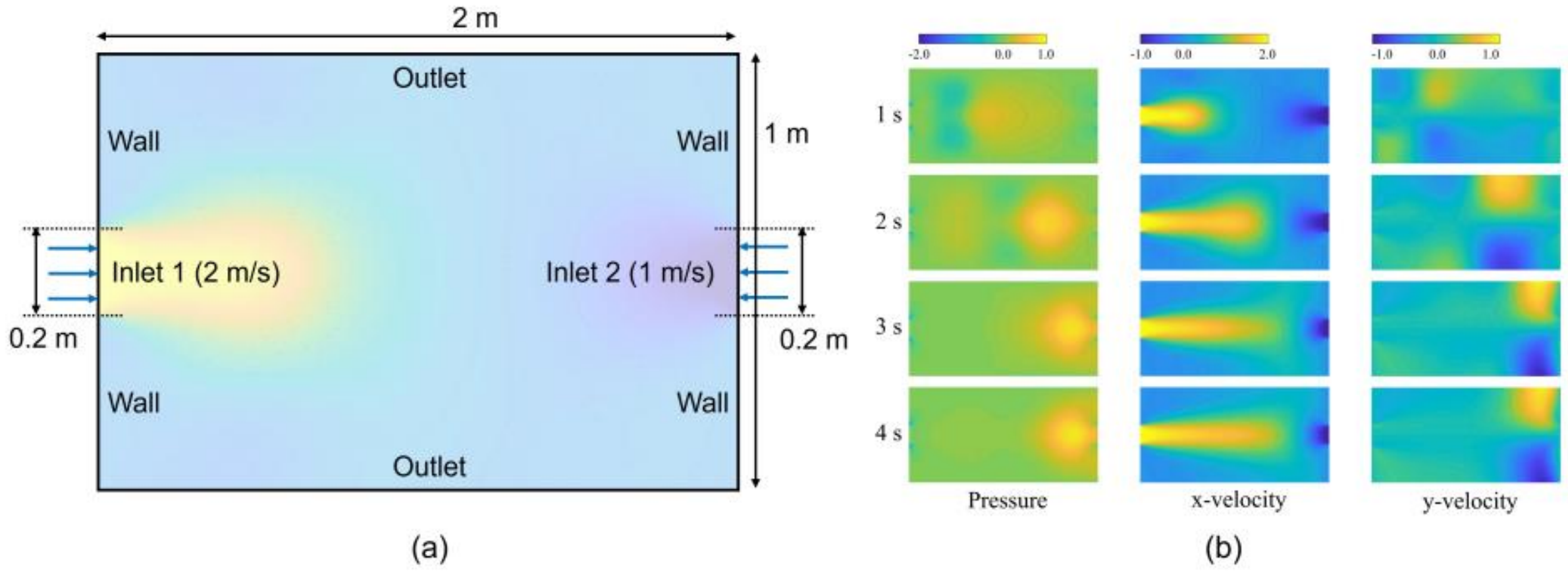

**Figure 11**. Laminar flow CFD dataset from the counterflow simulation. (a) computational domain and boundary conditions with nonslip walls. (b) time series until formation of stabilized counterflow. The most unsteady 1.00-1.11 s timeseries dataset were used for network performance evaluation.

5.2. Efficacy of physics-informed loss function

**Fig. 12** shows the change in FVMN network design with a physics-informed loss function. The loss function in the previous network (**Fig. 12(a))** is simply calculated as the MSE between the predicted value and ground truth value as shown **Eq. (52)**. Where the number of grids $n$ was 18,000. 80% of the total grids (14,400) was used for the training loss and 20% (3,600) was used for the validation loss for checkpoints.

$$L = L_{mse} = \frac{1}{n}\sum_{k=1}^{n}\left(Z^k - Z^k(\theta)\right)^2 \tag{52}$$

On the contrary, an algorithm for the physics-informed loss function was developed as shown **Eq. (53)**. The residual $\varepsilon_j^k$ for each grid $k$ and each governing equation $j$ is iteratively calculated for the parameters determined in the mini-batch and included in the loss function. As shown **Fig. 12(b)**, the residual is calculated by summation of physical fluxes of each governing equation, continuity and the Navier-Stokes equation (x-momentum and y-momentum). Because all variables $(p, u, v)$ were predicted in the unified network to compute the residuals during network training, the MSE loss function also summed squared error of each grid $k$ and each variable $i$. $w_1$ and $w_2$ is weighting factor to adjust the scale of each function ($w_1 = 1, w_2 = 0.5$ in this study). These factors of hyperparameter properties can prevent ignoring the effect of some losses due to scale differences during network training [32, 55]. Additionally, the continuity residual was scaled by 1/10 to consider the residual of the continuity and the Navier-Stokes equation together. The residuals of 100 % grids were averaged and used for training loss $L_{residual}$. Because $L_{residual}$ calculates the loss without the ground truth value, it can be classified in an unsupervised manner. Although the total loss function including $L_{mse}$ has a supervised manner, $L_{residual}$ can be utilized as an unsupervised monitoring function in the prediction dataset. In MSE function, same as network (a) 80% of the total grids was used for the training loss and 20% was used for the validation loss for checkpoints. **Table 3** shows the summary of each network design based on input/output system and loss function; general MLP model, FVMN model and FVMN model with the physics-informed loss function (PI-FVMN).

$$L = w_1 L_{mse} + w_2 L_{residual} = w_1 \cdot \frac{1}{n} \sum_{i=1}^{3} \sum_{k=1}^{n} \left( Z_i^k - Z_i^k(\theta) \right)^2 + w_2 \cdot \frac{1}{n} \sum_{j=1}^{2} \sum_{k=1}^{n} \left( \varepsilon_j^k \right)^2 \quad (53)$$

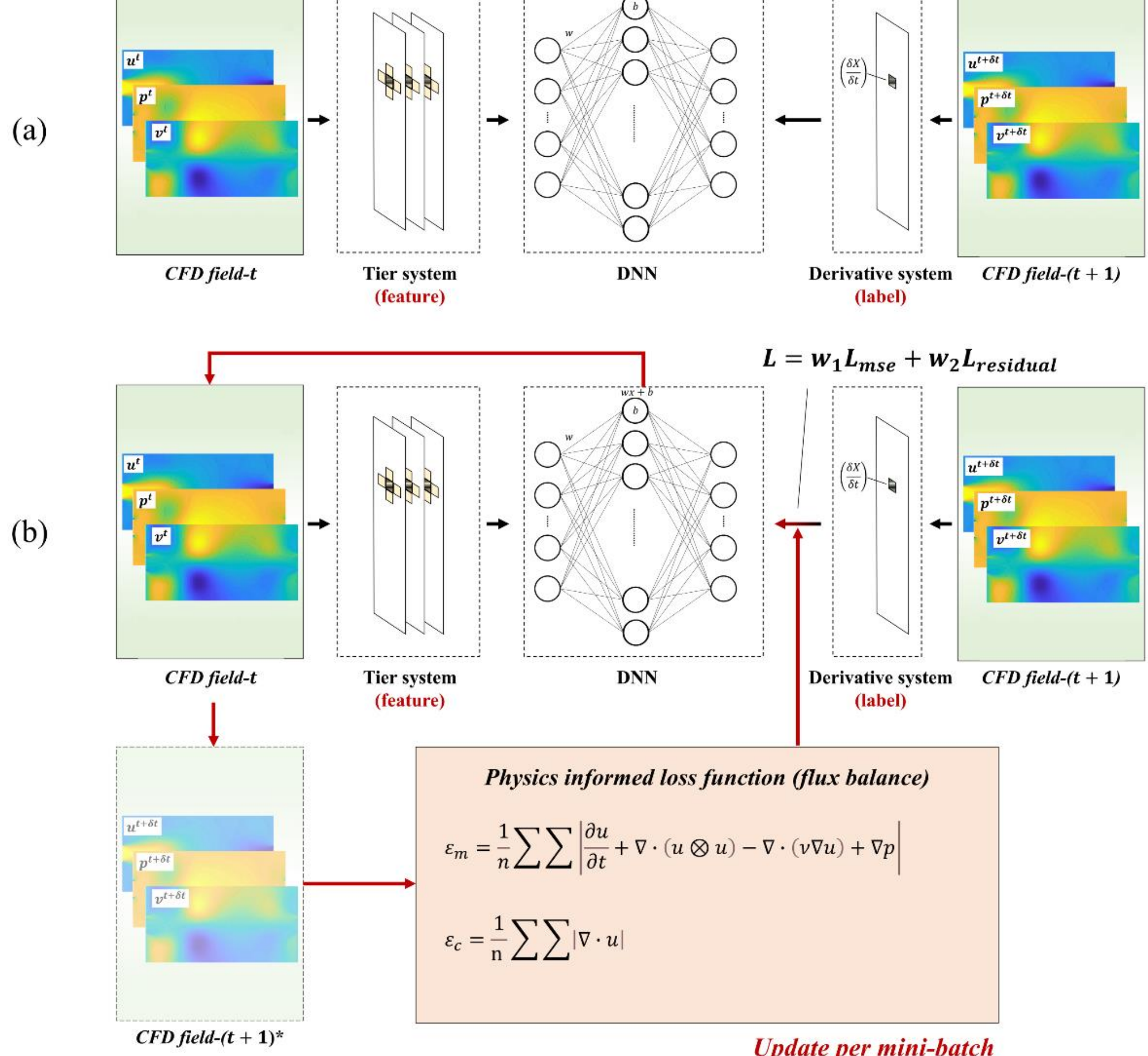

**Figure 12**. FVM network design with (a) MSE loss function. (b) physics-informed loss function. The physics-informed loss function updates the sum of the MSE and residual of conservation equations per mini-batch.

**Table 3.** A summary of each network design

| Network type | Feature ($t$) | Label ($t+1$) | Loss function |
|---|---|---|---|
| **General MLP** | $X^t$ –Variables in main grid | $X^{t+1}$ –Each variable in main grid | $L_{MSE}$ |
| **(a) FVMN** | $X^t_{i,j}$ –Variables in main/neighboring grids | $(X^{t+1} - X^t)$ –Each variable derivative in main grid | $L_{MSE}$ |
| **(b) PI-FVMN** | $X^t_{i,j}$ –Variables in main/neighboring grids | $(X^{t+1} - X^t)$ –Each variable derivative in main grid | $L_{MSE} + L_{residual}$ |

We investigated the efficacy of the physics-informed loss function by comparing the performance of network (a) and (b). As in the reacting flow simulation, single time series was used for network training and 10 time series data were used for performance evaluation. More specifically, the training/validation dataset was 80/20% of the 1.00–1.01 s time series data and test dataset was the 1.01-1.11 s time series data. The hyperparameters were optimized by identical manner in *Section 4*: 2 hidden layers, 64 node/layer, mean square error (MSE) and ReLU function. The training process was terminated when the loss function converged. It seems reasonable to optimize to a fewer hidden layers than previous reacting flow dataset generated from relatively more complex governing equations.

**Fig. 13(1)** shows the effect of the physics-informed loss function in the training process. Network (a) and (b) is the network with the MSE loss function and the physics-informed loss function, respectively. Network (a)-double training data is the identical network condition as network (a), but the training dataset size is doubled (1.00-1.02 s). In network (a), the training loss and total residual (sum of the 1/10 scaled residual of continuity and the residual of the Navier-Stokes equation) gradually converged from about 300 epochs. In case network (b) and network (a)-double, the convergence trend of losses was similar but the values of the converged total residuals were different.

**Fig. 13(2)** compares the converged value of each loss between cases. When the residual term was included in the loss function (network (b)), the total residual converged to a lower value without compromising the training/validation error. Intriguingly, the magnitude of the residual was even smaller than network(a)-double which was trained by the double dataset size. It means that providing the governing equation information in the loss function can increase network performance as much as increasing the amount of training data. Although the training/validation losses in the training dataset did not have significant difference in all cases, the reduced residual can affect the accuracy of the test dataset (1.01-1.11 s).

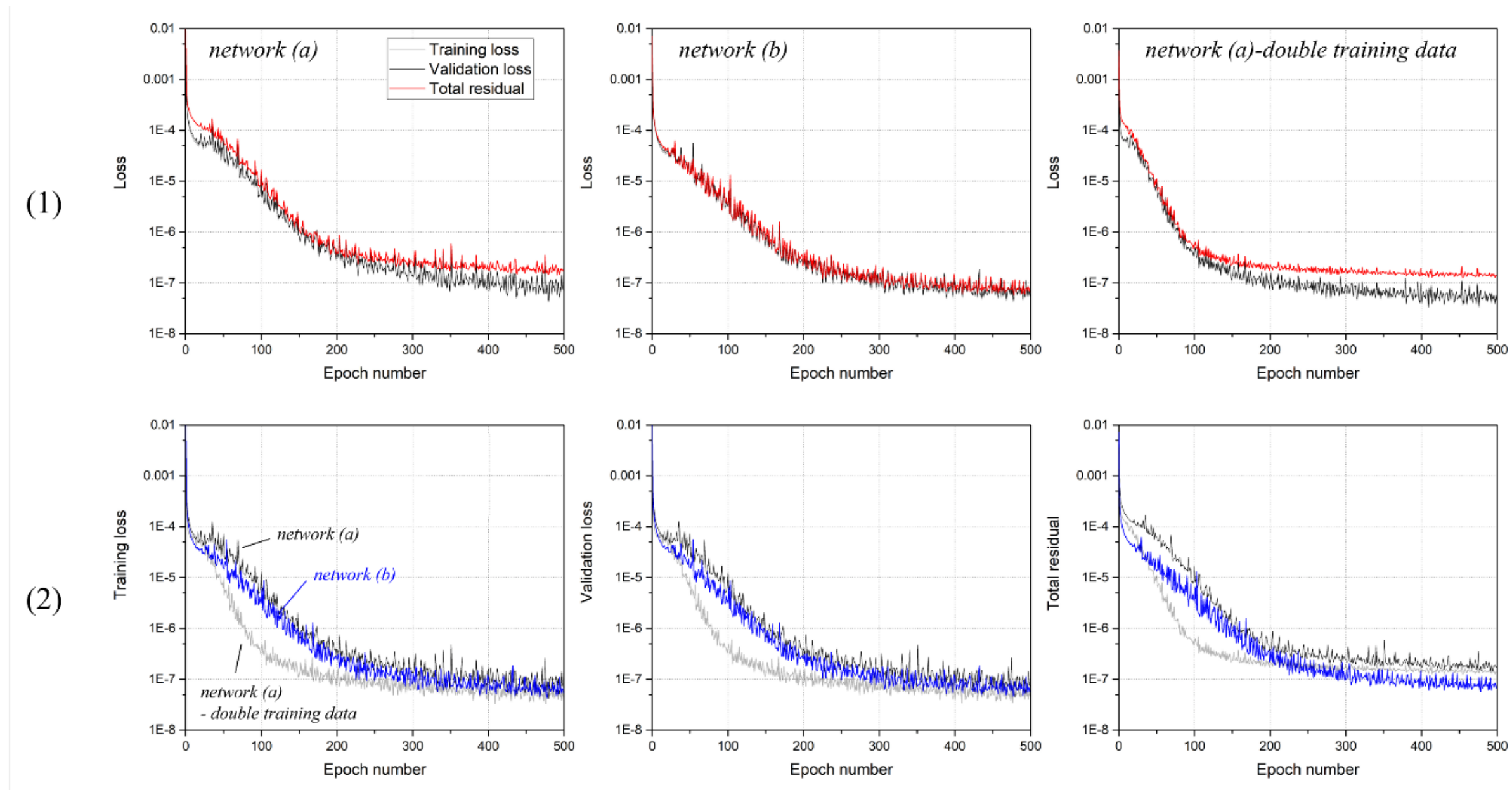

**Figure 13**. (1) variation of training/validation losses and total residual in network training (logscale). (2) comparison of variation on each loss according to the network model (logscale). When the residual term was included in the loss function (network (b)), the total residual converged to a lower value without compromising the training/validation error.

**Fig. 14(left)** shows the variation of network accuracy in the test dataset (1.01-1.11 s time series). Because the ground truth velocity values of some fields are close to zero, the network accuracy was compared based on the absolute error. First, the maximum absolute error of each variable gradually increased with timestep as similar in *Section 4*. In network (a), the $u$-maximum error of about 0.03 m/s in the last timestep was an error of 1% based on the maximum velocity in the domain. Considering that the network predicted a dataset 10 times the training dataset, the FVMN shows good agreement with the ground truth data. In network (b), the $u$-maximum error of about 0.015 m/s in the last timestep was about half of the error in network (a). Although the $p$-maximum error (Pa) increased as compared to network (a), the overall accuracy was improved by the physics-informed loss function.

**Fig. 14(right)** shows the variation of residuals of the governing equations such as continuity and Navier-Stokes equation. The residuals can be calculated in an unsupervised manner without a ground truth data. Interestingly, the trained network (b) considering the physical flux balance in training process always showed a lower residual value than network (a) for the test dataset. Especially, the residual in the Navier-Stokes equation (momentum balance equation) where all variables are included in the equation showed a significant difference. We concluded that the physics-loss function can prevent the non-physical overfitting problem where the network is trained to reduce the difference between the predicted value and ground truth while ignoring the physical flux balance. For this reason, the $p$-maximum error in network (b) was slightly larger than the u-maximum error in network (a), but it is reversed by a large difference (**Fig. 14(left)**). In general CFD simulations, an elevated residual value can reduce the reliability of simulation results. The reduction of the residuals contributed to the improvement of the multistep time series prediction accuracy, but it seems not a remarkable improvement due to sensitivity for each variable.

More importantly, in both networks the residuals increased with time and the maximum error increased accordingly. Without CFD ground truth, we can estimate the network accuracy trend over timestep by calculating the residual in an unsupervised manner. Additionally, CFD field prediction for longer time series may be feasible if we can prevent the increase of the residual with timestep. This applicability of the physics-informed loss function provides the feasibility of the ML-CFD cross coupling strategy in the next section. Consequently, two intriguing scientific observations are clarified in this section: (1) the physic-informed loss function can prevent non-physical overfitting problem and ultimately reduces the error in test dataset (2) observing the calculated residuals in an unsupervised manner can indirectly estimate the network accuracy.

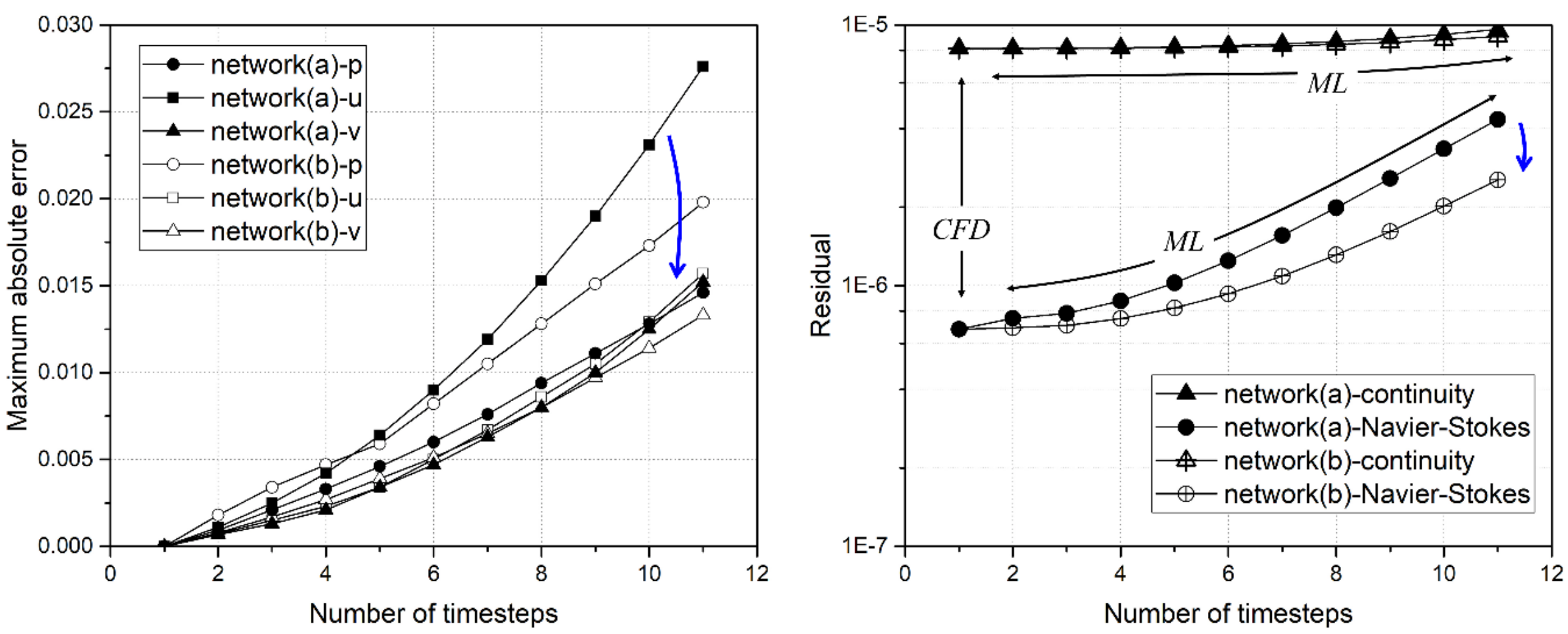

**Figure 14**. Variation of maximum absolute error of each variable (left) and calculated residual of each governing equation (continuity and the Navier-Stokes equation) in ML prediction time series (right, logscale).

5.3. Concept of ML-CFD cross coupling framework

In previous sections, it was confirmed that the accuracy of the FVMN was significantly improved compared to the general MLP model. In the reacting flow dataset, the network trained with single CFD time series data showed a mean relative error of 0.24% in the temperature field for the $10^{th}$ ML prediction time series data. In counterflow dataset, it was identified that the physics-informed loss function prevented the non-physical overfitting problem. Notably, the improved accuracy of the FVMN can be attributed to the proposed algorithm which thoroughly projects the CFD principles in network architecture and loss function. Although this improved network performance for prediction dataset (test dataset) is very encouraging, the linearly increasing gradient error is still a remaining issue when transient calculation times are longer. In other words, improvements in the baseline model can reduce the error increase rate, but the linear error increase is seemingly inevitable in ML approaches. Recently, Raissi et al. suggested that unsteady 1D PDE calculation is feasible by including residuals of some random grid in future time series [54]. However, including residuals in future time series is not suitable for CFD computation acceleration purpose.

Therefore, in this study, we suggest the concept of ML-CFD cross-coupling framework, which is an ML-aided CFD computation that can accelerate the unsteady heat transfer and flow field simulation (**Fig. 15**). The key idea is that CFD field prediction for longer time series will be practical if we can prevent the increase of the residual with timestep. Initially, the CFD calculation for a single step is conducted by solving the first principles. The number of timesteps included in the training dataset may depend on the complexity of each simulation. In the next step, the calculated CFD time series data train the constructed neural networks to determine the parameters. The trained networks can then quickly predict the multistep CFD data without costly simulation. Up to this point, the foregoing prediction process is identical to the multistep prediction case discussed in the previous section. However, the computation framework reverts to the CFD calculation when the ML computation error approaches the tolerance. This method seems feasible because an acceptable residual range exists even in the CFD simulation.

When an ML time series is input as the CFD initial condition and the next time series is calculated by the first principle, the residuals are reduced to well below the CFD tolerance level through internal iterations. If the residuals of the input ML time series did not exceed the CFD tolerance level, error divergence can be prevented as in general CFD simulations. Them the new CFD results are applied to the training dataset to update the neural network parameters anew. Of note, the intermediate parameter updating time is much shorter compared to the initial network training time. As shown **Fig. S9**, the losses were stabilized with only 2 epochs in contrast to the initial training which required about 500 epochs in counterflow dataset. The evolved neural networks again accelerate the CFD simulation by calculating the multistep data; this process is repeated until the calculation is completed. The key advantage of using the computation framework is that it can be applied individually to each simulation. Although some amount of network optimization is required depending on the simulation complexity, network training using numerous simulation results is unnecessary.

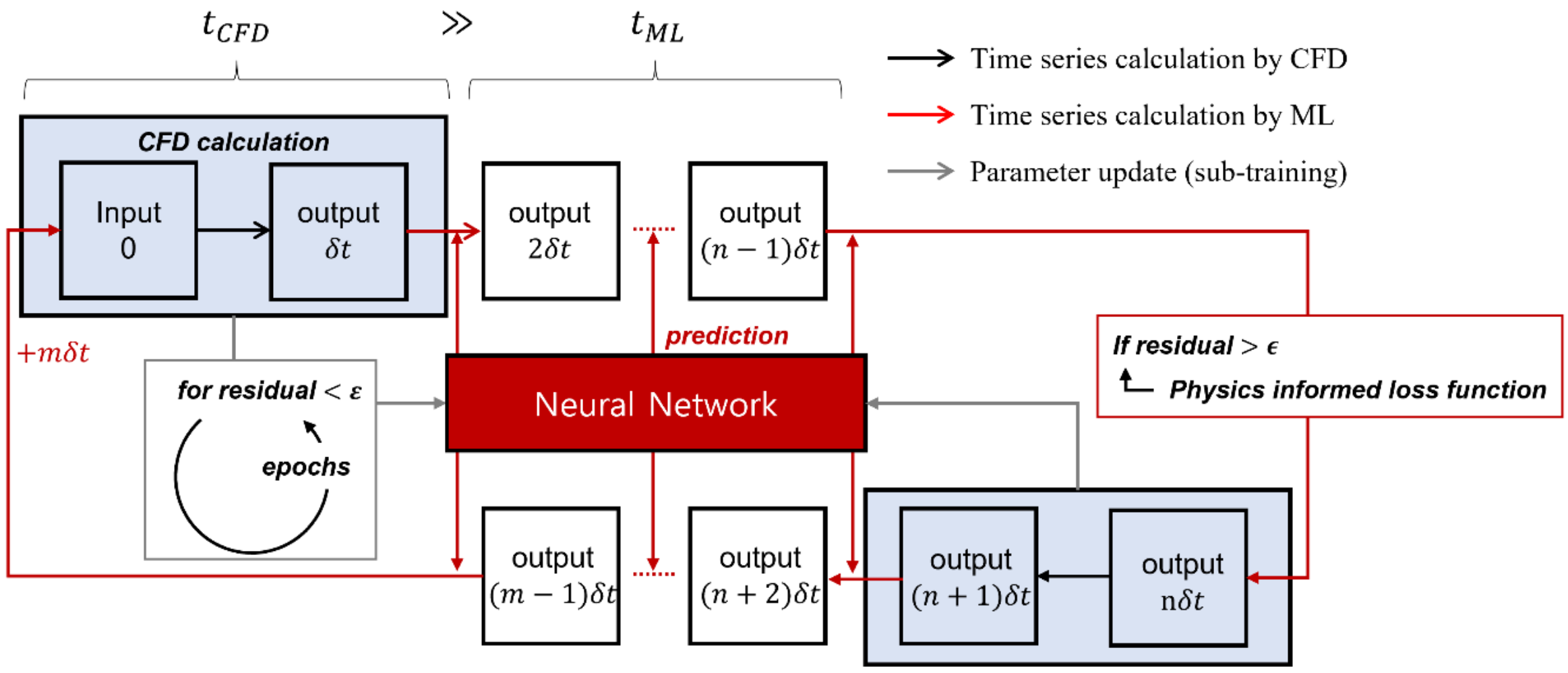

**Figure 15.** Concept of the ML-CFD cross coupling framework accompanied by periodic parameter updates through the physics-informed loss function. Notably. the computation time of machine learning for one timestep $t_{ML}$ is much smaller than that of CFD $t_{CFD}$ (about 10 times faster in the reacting flow dataset).

For the ML-CFD framework to accurately predict CFD time series over a long period, the trend of error variation in each ML timestep must be appropriately monitored. In predicting time series data, ground truth data (i.e., CFD results) are missing unlike in the training dataset. This means that the relative error compared with ground truth data cannot be periodically checked to ascertain whether tolerance has been exceeded. In view of this, we considered the use of the residual function as a monitoring function. As confirmed in the previous section, the residual of ML prediction time series calculated in unsupervised manner can estimate the variation of the network accuracy.

As in *Section 5.2*, the continuity residual of the ML prediction time series $\varepsilon_c^t$ was also calculated in the reacting flow case as shown **Fig. 8**. As shown **Eq. (54)**, the residual value was scaled through reference residual values in the training dataset $\varepsilon_c^*$ which matches the tolerance level of the CFD simulation. In the ANSYS-Fluent code, the residual was scaled by the dominator with the largest absolute value in the first five iterations [48]. As discrepancies between the input variables of training and prediction datasets increased, the mass conservation error in the trained network gradually increased. Consequently, the feasibility of estimating the network accuracy through the residual monitoring seems independent of the type of CFD simulation.

$$\varepsilon_{c,scaled}^t = \frac{\varepsilon_c^t}{\varepsilon_c^*} \tag{54}$$

As a result, we confirmed that the degradation of neural networks can be estimated by residual calculations without the ground truth of flow fields. It is worth to note that our current approach has a limitation that heat transfer and flow field prediction by the network is valid under the same geometry condition of the training dataset (same mesh). This is because, unlike the CFD numerical scheme, the change of the physical flux balance coefficient according to the mesh construction was not considered in this study. To predict the CFD time series by including the mesh size as input, a super deep learning network should be implemented by enormously increasing the training data size and network size. This approach is deemed impractical, and it again requires an enormous training time for even small changes in a governing equation of a simulation.

On the contrary, our approach is expected to be very useful for the CFD simulation requiring a long CPU time. As such, the network trained by the initial simulation results at each simulation condition can effectively accelerate the CFD simulation. As discussed in *Section 1*, recent benchmark studies show that approximately 10–100 h of CPU time per second of physical time is required to simulate unsteady hydrogen deflagration in the compartment unit [4]. If the suggested ML-CFD cross-coupling framework becomes more definitive through various case studies, then it may be a potential solution for overcoming the limitations of complex CFD simulations. Although full acceleration of CFD simulation is difficult with this teacher-forcing approach, this cross-coupling framework (partial acceleration) seems to be an effective application method of data-driven modeling. The concrete verification and acceleration performance evaluation of this framework through a case study is our future work.

## 6. Conclusion

In this study, a new network model was developed to predict multistep CFD time series data more accurately. The developed model introduced the FVM principle, which is the basic numerical method of most CFD codes, with a unique network architecture and physics-informed loss function. The developed network model, considering the nature of the CFD flow field where the identical governing equations are applied to all grids, can predict the future fields with only two previous field**s**, unlike the CNN**s** requiring a large amount of field images. We evaluated the individual efficacy of tier-input and derivative-output systems based on an unsteady reacting flow simulation with 20 detailed chemical reactions. The FVMN model was confirmed to be capable of accurately predicting the CFD time series of a prediction dataset 10 times the size of the training dataset. The scale separation of output variables through the derivative system significantly improved network performance. As the next step, we compared the performance of the FVMN model with the MSE loss function and the model with the physics-informed loss function. Consequently, we identified two intriguing scientific observations: (1) the physic-informed loss function can prevent non-physical overfitting problem and ultimately reduce the error in test dataset (2) observing the calculated residuals in an unsupervised manner can indirectly estimate the network accuracy.

Although the relative error with respect to ground truth data was significantly reduced by the developed network model, the linearly increasing gradient error remained a problem in long transient calculation times. Hence, we suggested the concept of ML-CFD cross-coupling framework, which is an ML-aided CFD computation that can accelerate the unsteady heat transfer and flow field simulation. Without CFD ground truth, we can estimate the network accuracy trend over timestep by calculating the residual in an unsupervised manner. This applicability of the physics-informed loss function provides the feasibility of the ML-CFD cross coupling strategy. The concrete verification of this framework through a case study is our future work.

**Acknowledgements**

This work was supported by the Nuclear Safety Research Program through the Korea Foundation of Nuclear Safety (KoFONS) using the financial resource granted by the Nuclear Safety and Security Commission (NSSC) of the Republic of Korea (Grant No. 2003006-0120-CG100).

[**Supplementary material**]

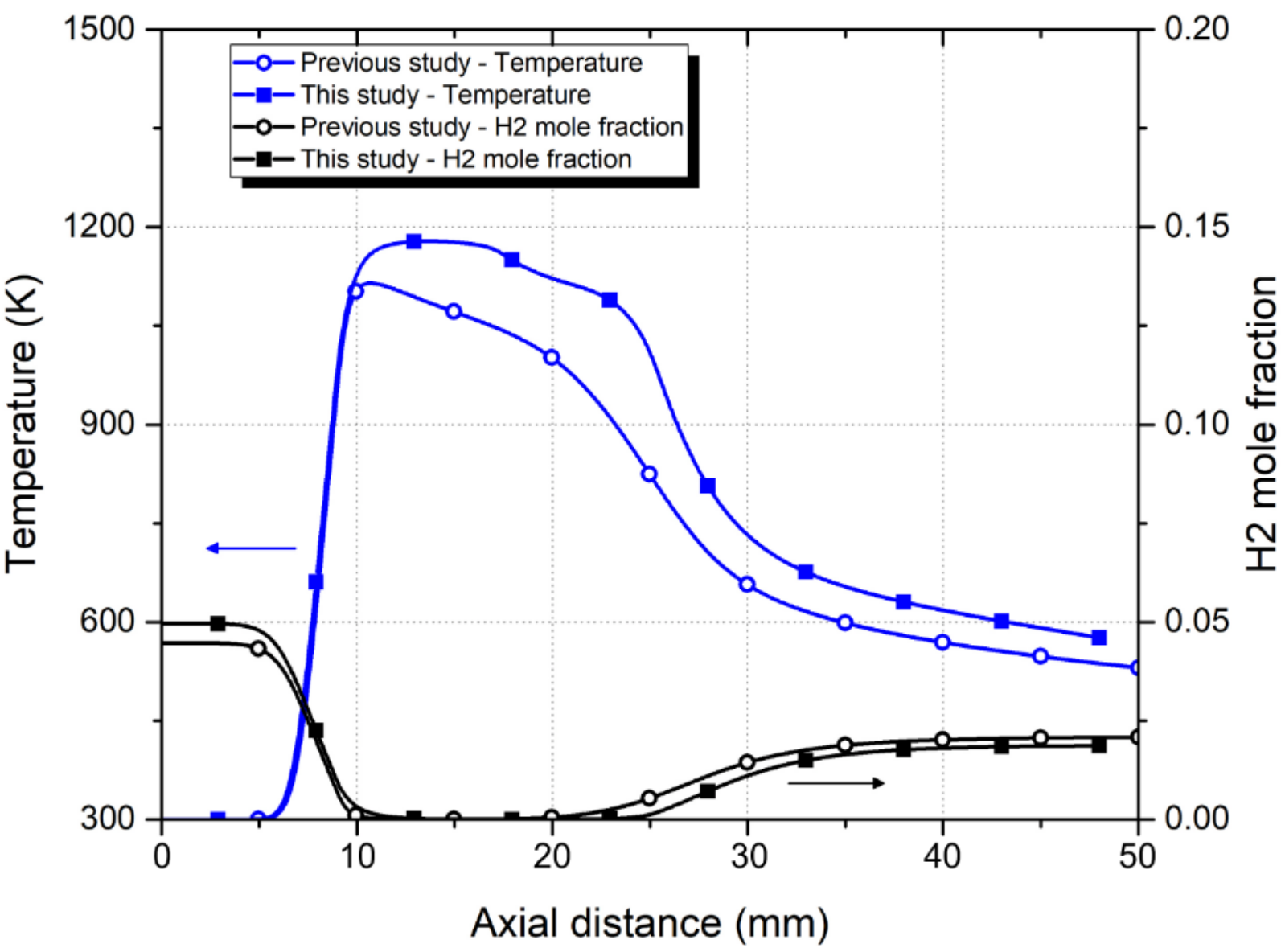

**Figure S1.** Verification of the transient simulation results through comparison with the previous steady results. Since the initial hydrogen concentration was as high as 0.5% in this study, the maximum temperature also increased slightly.

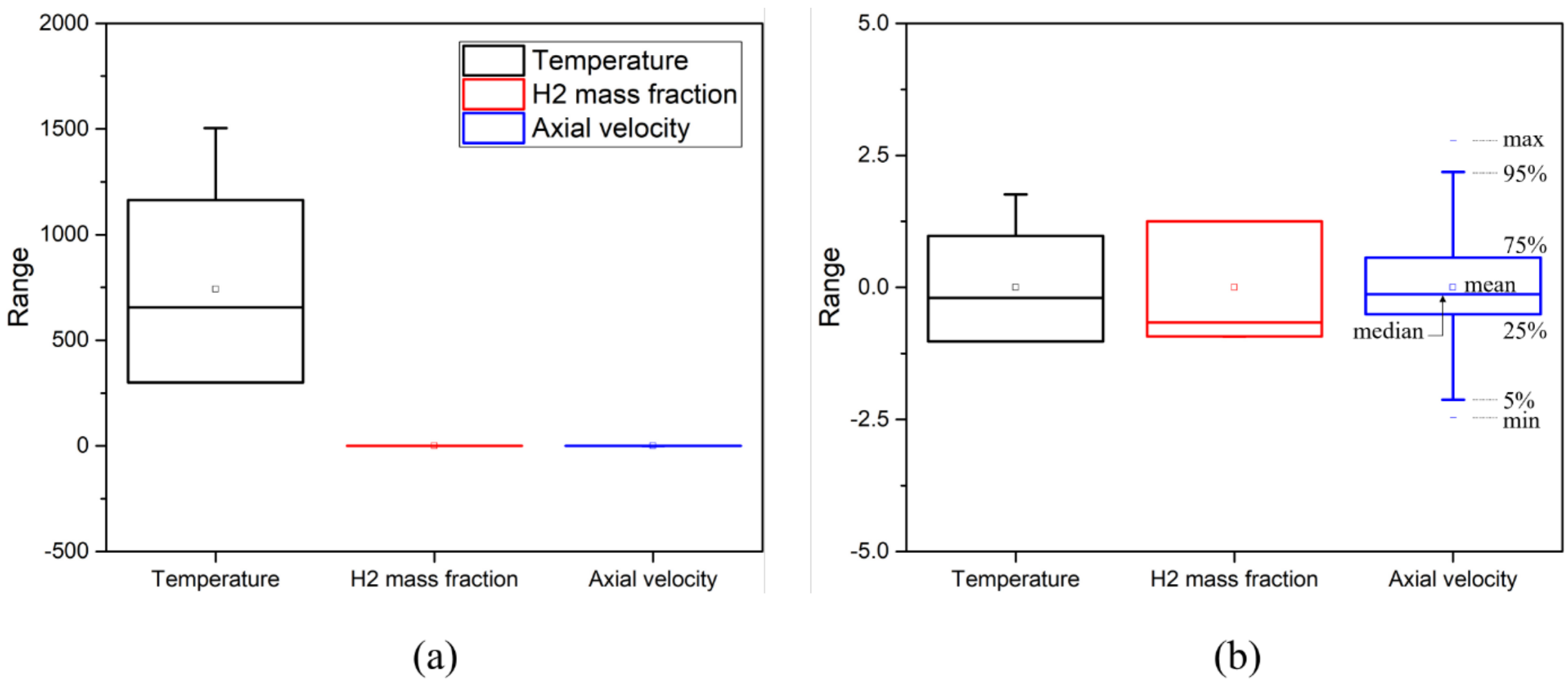

**Figure S2.** Boxplots of input variables (a) before and (b) after standardization; apparent difference in variable scale is resolved.

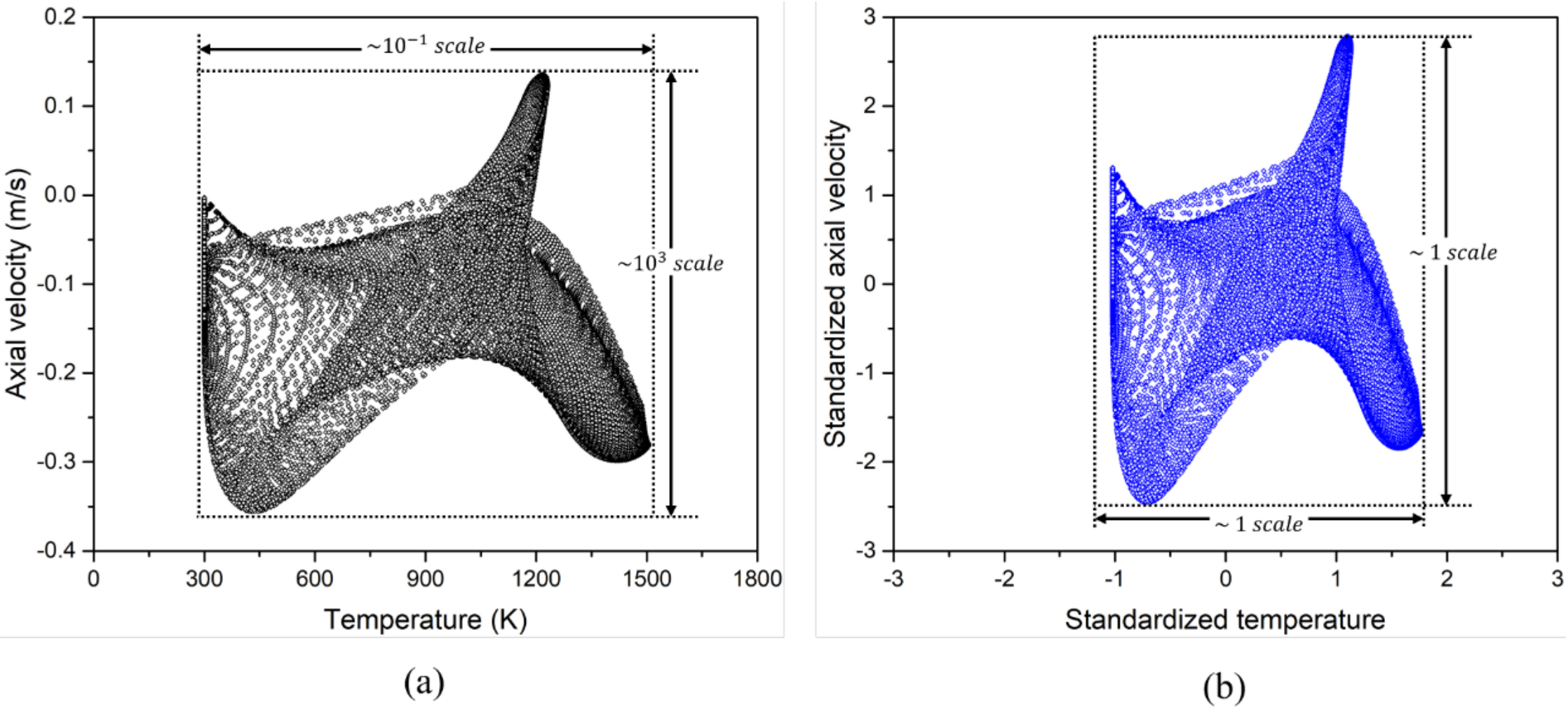

**Figure S3.** Before (a) and after (b) standardization of features to resolve large differences between their ranges. It was confirmed that the features including temperature and axical velocity were transformed to comparable scales.

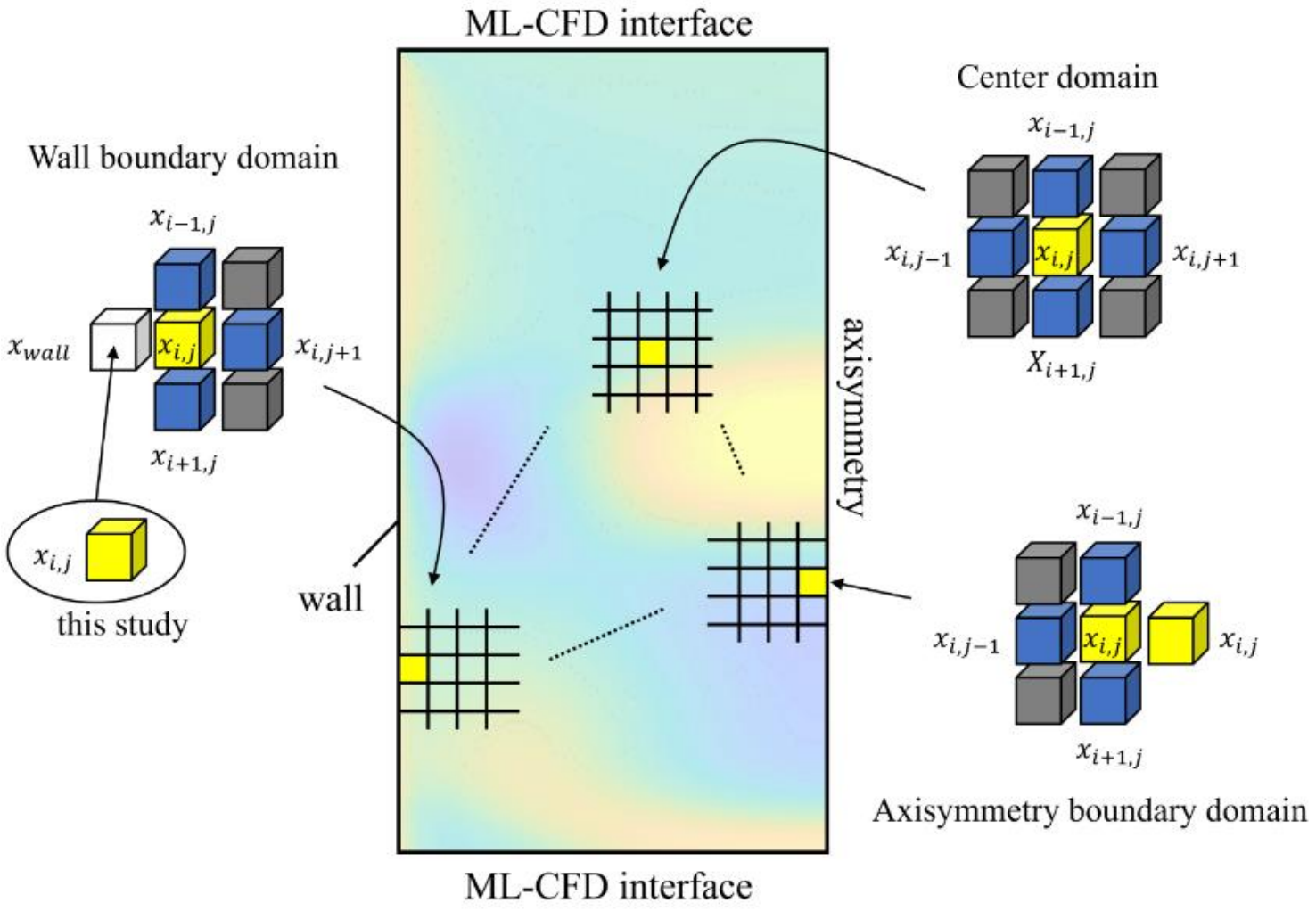

**Figure S4**. Schematic of ML input datasets in boundary domain for tier system.

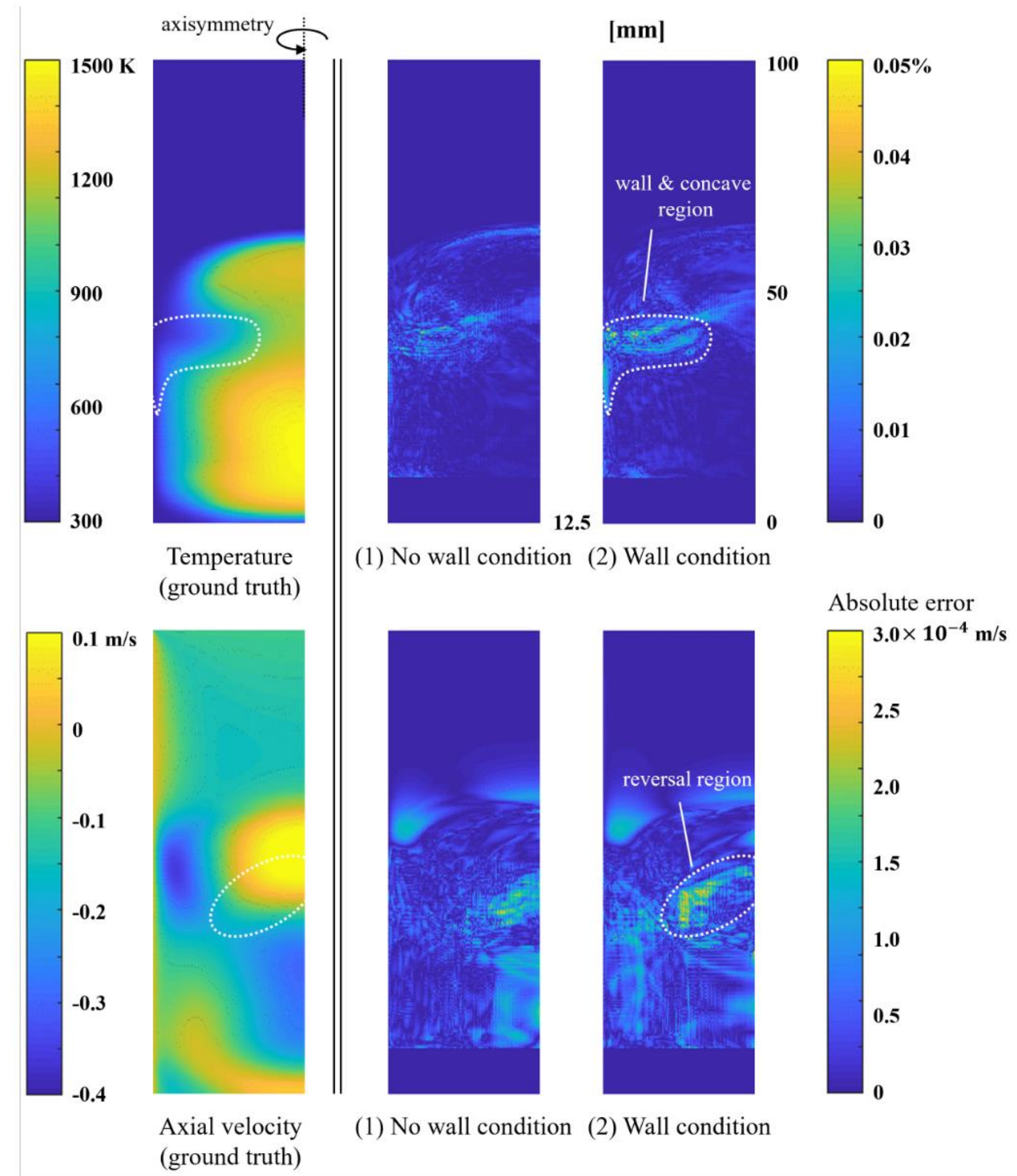

**Figure S5**. Error fields of predicted temperature $\frac{|T_{ML}-T_{CFD}|}{T_{CFD}}$ and axial velocity $|u_{ML} - u_{CFD}|$ according to the presence of the wall condition. The error in the gradient region of each variable became more pronounced at the wall condition.

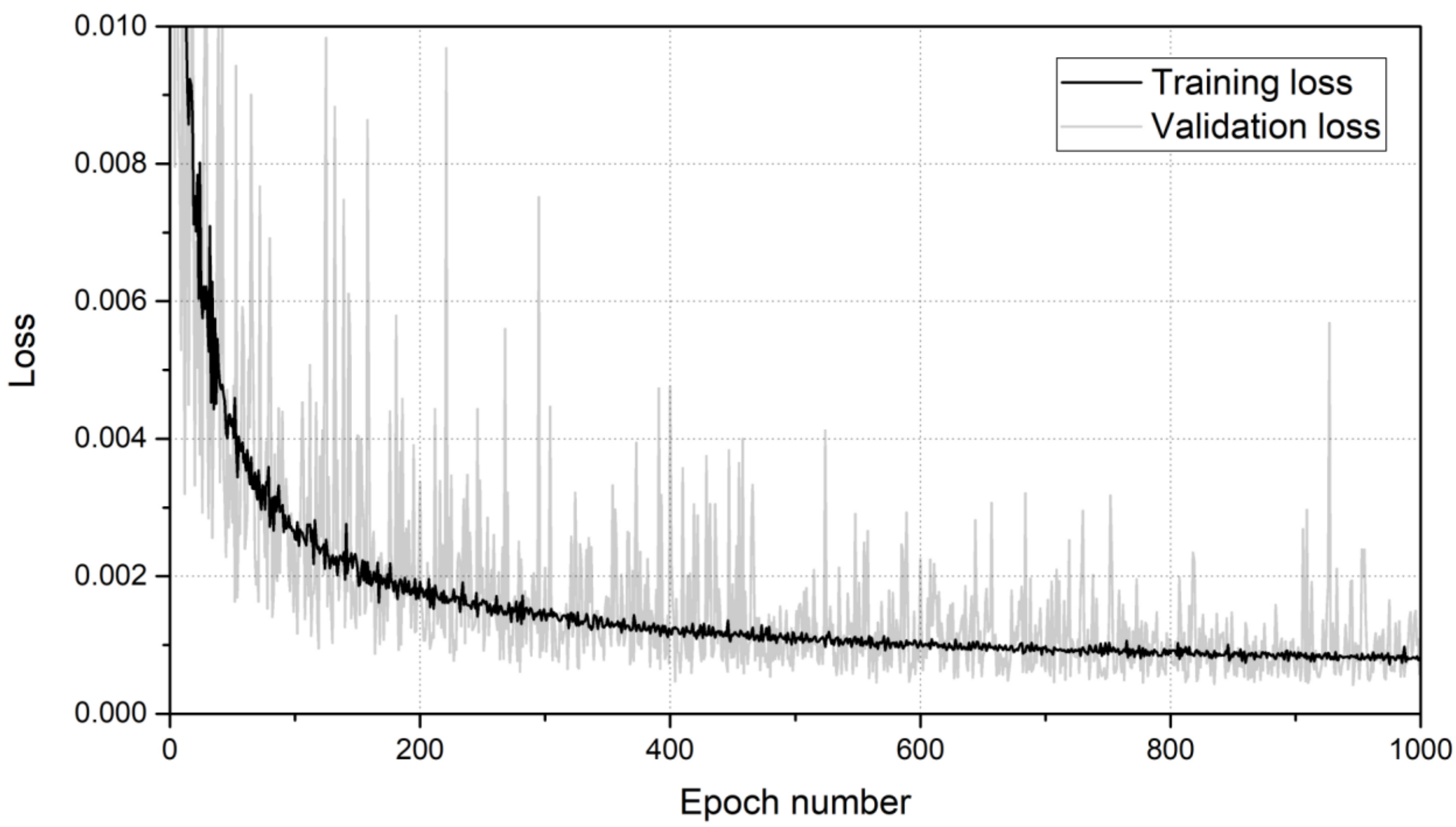

**Figure S6**. Variation of training and validation loss in network training in case (c). As the validation loss oscillates around the training loss value, both losses were gradually converged after 500 epochs. By the loss variation, it was confirmed that the amount of data was sufficient to prevent overfitting.

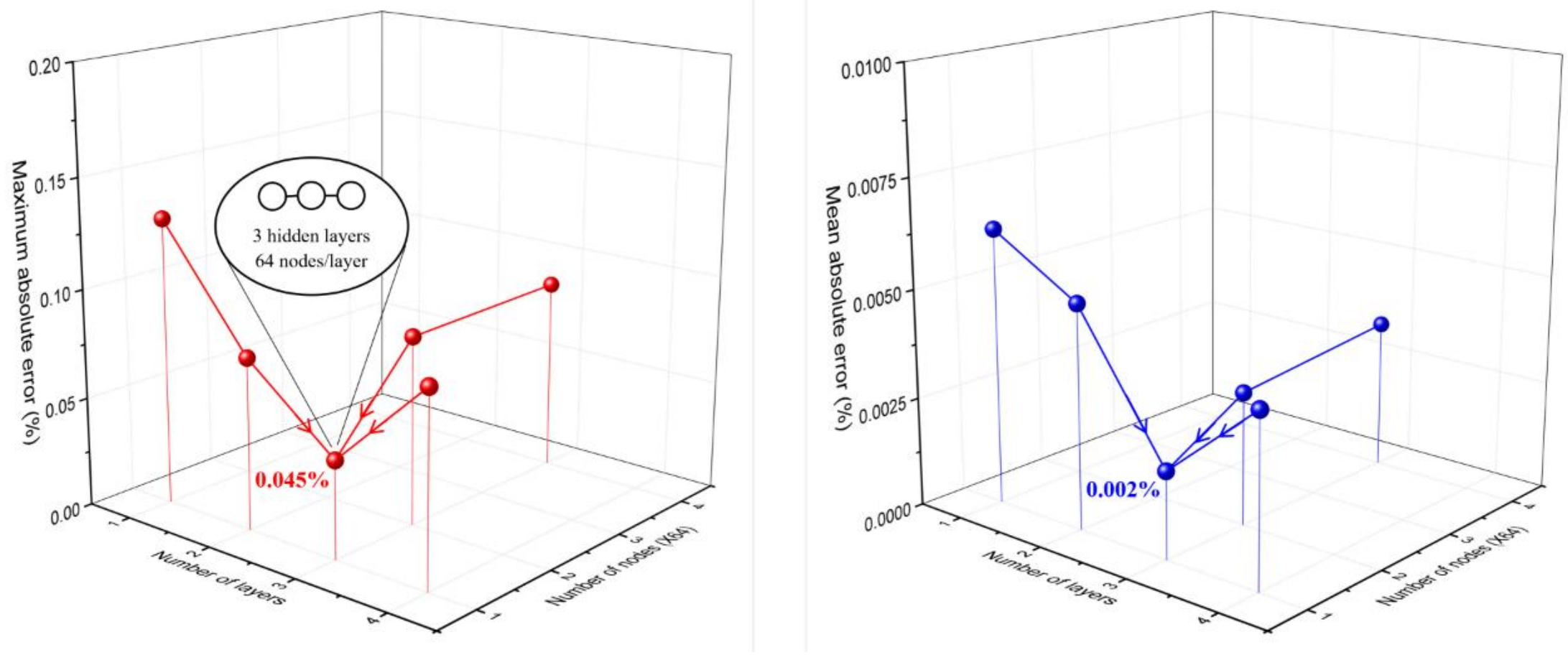

**Figure S7**. Changes in maximum absolute error (left) and mean absolute error (right) according to the number of hidden layers and nodes.

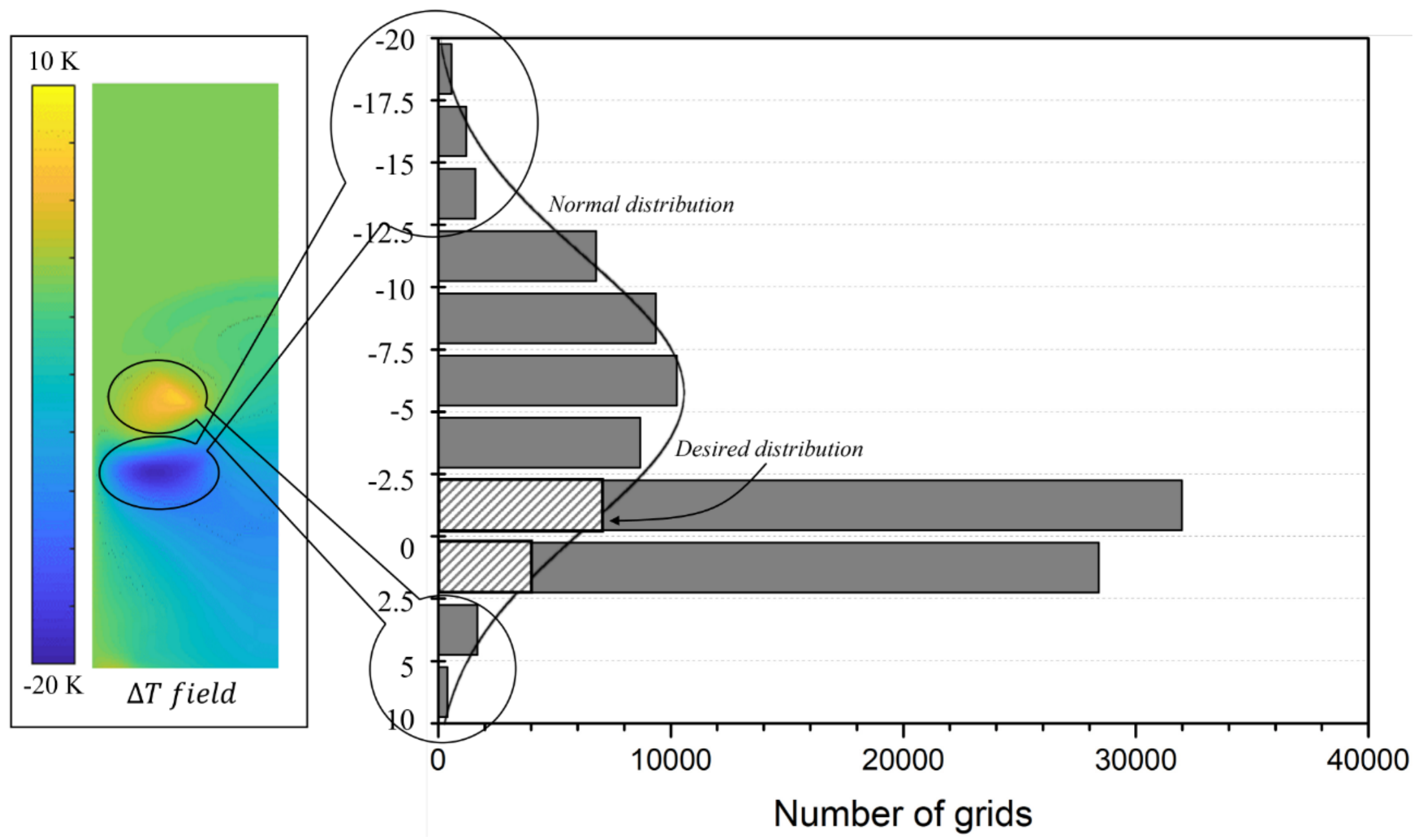

**Figure S8**. Distribution of output variable, $Z_{i,d}^{t+1} = \left[\left(\frac{\delta x_i}{\delta t}\right)_{i,j}^{t+1}\right]$, in CFD temperature field at 0.600–0.601 s; biased data distribution may cause considerable error in stiff region.

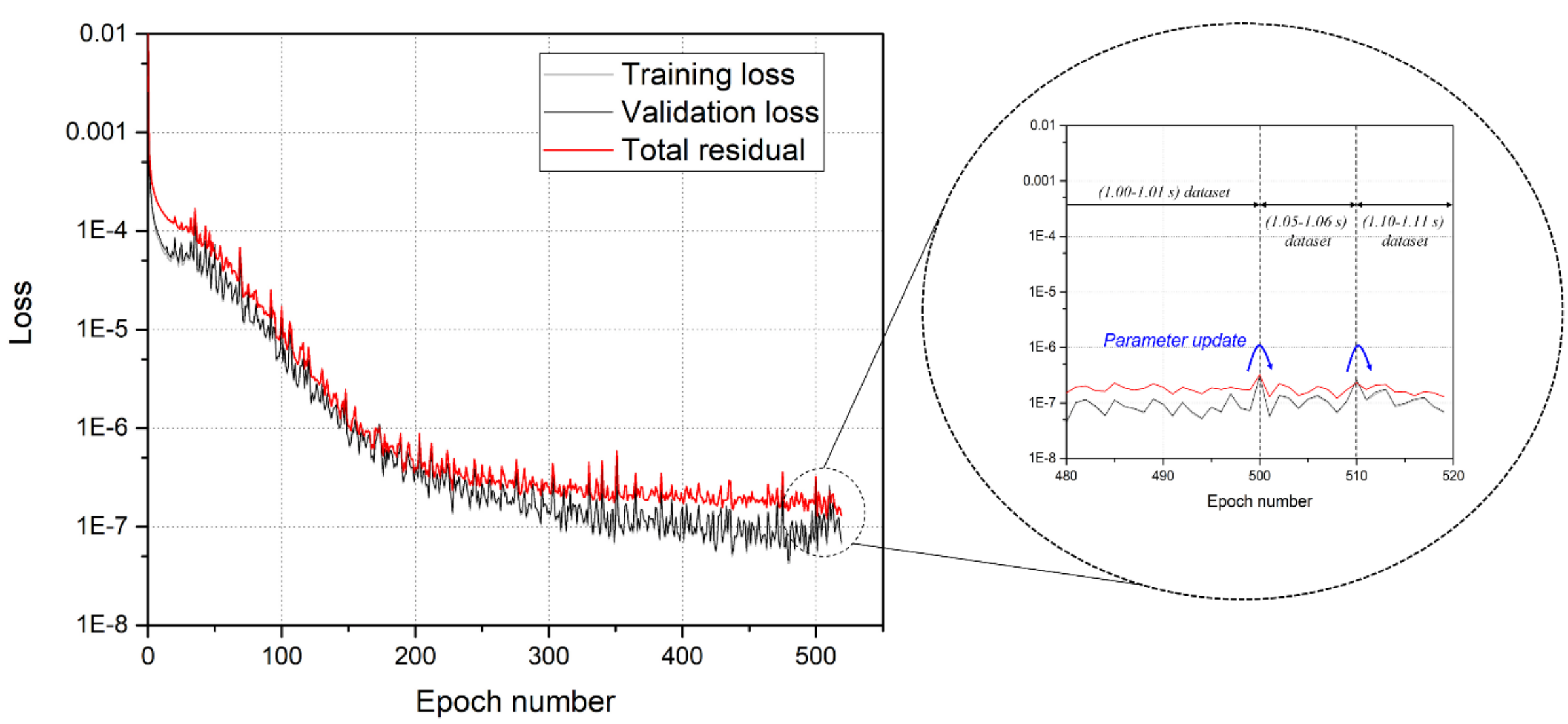

**Figure S9**. Variation of training/validation losses and total residual in network training and updating (training: 1.00-1.01 s dataset, updating: 1.05-1.06 s dataset and 1.10-1.11 s dataset). It was noted that the parameter updating with a much smaller epochs (~2 epoch) than initial training was feasible.